\PassOptionsToPackage{table,dvipsnames}{xcolor}
\documentclass{nashnova-paper}
\usepackage{xcolor}
\usepackage{amssymb}
\usepackage{amsmath}
\usepackage{amsthm}
\usepackage{bm}
\usepackage{multirow}
\usepackage{longtable}
\usepackage{needspace}
\usepackage{wrapfig}
\usepackage{float}
\let\sourcefigure\figure
\let\endsourcefigure\endfigure
\let\sourcetable\table
\let\endsourcetable\endtable
\renewenvironment{figure}[1][]{\sourcefigure[H]}{\endsourcefigure}
\renewenvironment{table}[1][]{\sourcetable[H]}{\endsourcetable}
\renewenvironment{figure*}[1][]{\sourcefigure[H]}{\csname float@endH\endcsname}
\renewenvironment{table*}[1][]{\sourcetable[H]}{\csname float@endH\endcsname}

\usepackage{enumitem}
\usepackage{subcaption}
\usepackage{makecell}
\usepackage{adjustbox}
\usepackage{threeparttable}
\usepackage{booktabs}
\usepackage{array}
\usepackage{tabularx}
\usepackage{pifont}
\usepackage{fontawesome5}
\usepackage{xspace}
\usepackage{graphicx}
\usepackage{colortbl}
\usepackage{tikz}
\usetikzlibrary{positioning,arrows.meta,fit,calc,shapes.misc,backgrounds}

\newcolumntype{C}[1]{>{\centering\arraybackslash}m{#1}}
\newcolumntype{Y}{>{\centering\arraybackslash}X}
\newcommand{\cmark}{\textcolor{nnPurple}{\ding{51}}}
\newcommand{\xmark}{\textcolor{gray}{\ding{55}}}
\newcommand{\pmark}{\textcolor{nnMuted}{$\boldsymbol{\sim}$}}
\newcommand{\paratitle}[1]{\vspace{1.0ex}\noindent\textbf{#1}\quad}

\theoremstyle{definition}

\newcommand{\bench}{\textsc{AgenticFinBench}\xspace}

\definecolor{cGround}{HTML}{2a78d6}
\definecolor{cVU}{HTML}{1baf7a}
\definecolor{cUnv}{HTML}{eda100}
\definecolor{cHal}{HTML}{e34948}
\definecolor{fpanel}{HTML}{FAF9FC}
\colorlet{fpanelb}{nnBorder}
\colorlet{fnavy}{nnDark}
\colorlet{fbrand}{nnPurple}

\title{
AgenticFinBench:
Benchmarking Financial Agents on Diverse Tasks from Real-World Requests
}

\author{NashNova Research}

\abstract{
Financial agents are becoming an important class of real-world AI systems. Financial tasks vary widely in type and difficulty, and many are open-ended with no ground-truth answers. However, existing benchmarks neither capture this breadth and open-endedness nor separate grounding from correctness. To bridge this gap, we introduce \textbf{\bench}, a benchmark of 230 real-world financial tasks spanning 8 categories, 2 languages, and 4 complexity levels.
We also propose a reference-free evaluation framework that combines question-specific rubrics with a two-stage claim audit that detects hallucinations, defined as claims absent from the agent's trace and refuted by external evidence. This framework better matches how financial work is assessed in practice.
Experiments across 18 agent configurations reveal three key findings:
(1) Ignoring hallucinations can overstate agent performance: hallucination auditing lowers the baseline leader's score by 7.58 points and moves it to fourth place.
(2) Harness differences explain more variation in evidence grounding than model differences (28.6\% vs.\ 8.8\%).
(3) Quality scores fall from 89.44 on direct retrieval to 79.18 on the most complex tasks. Case studies show that agents struggle to explain how events affect company profits or stock prices, compare financial data on a consistent basis, and justify their conclusions.
By releasing \bench, we aim to advance financial reasoning, factual verification, and agent design for more reliable responses to diverse real-world financial requests.
}

\date{October 29, 2026}
\gtechdata[\faHome\enspace Project Page]{https://nashnovaresearch.github.io/AgenticFinBench/}

\begin{document}
\maketitle

\vspace{-15pt}
\begingroup
\setlength{\intextsep}{0pt}
\begin{figure}[t]
  \centering
  \captionsetup{skip=5pt}
  \includegraphics[width=0.85\textwidth]{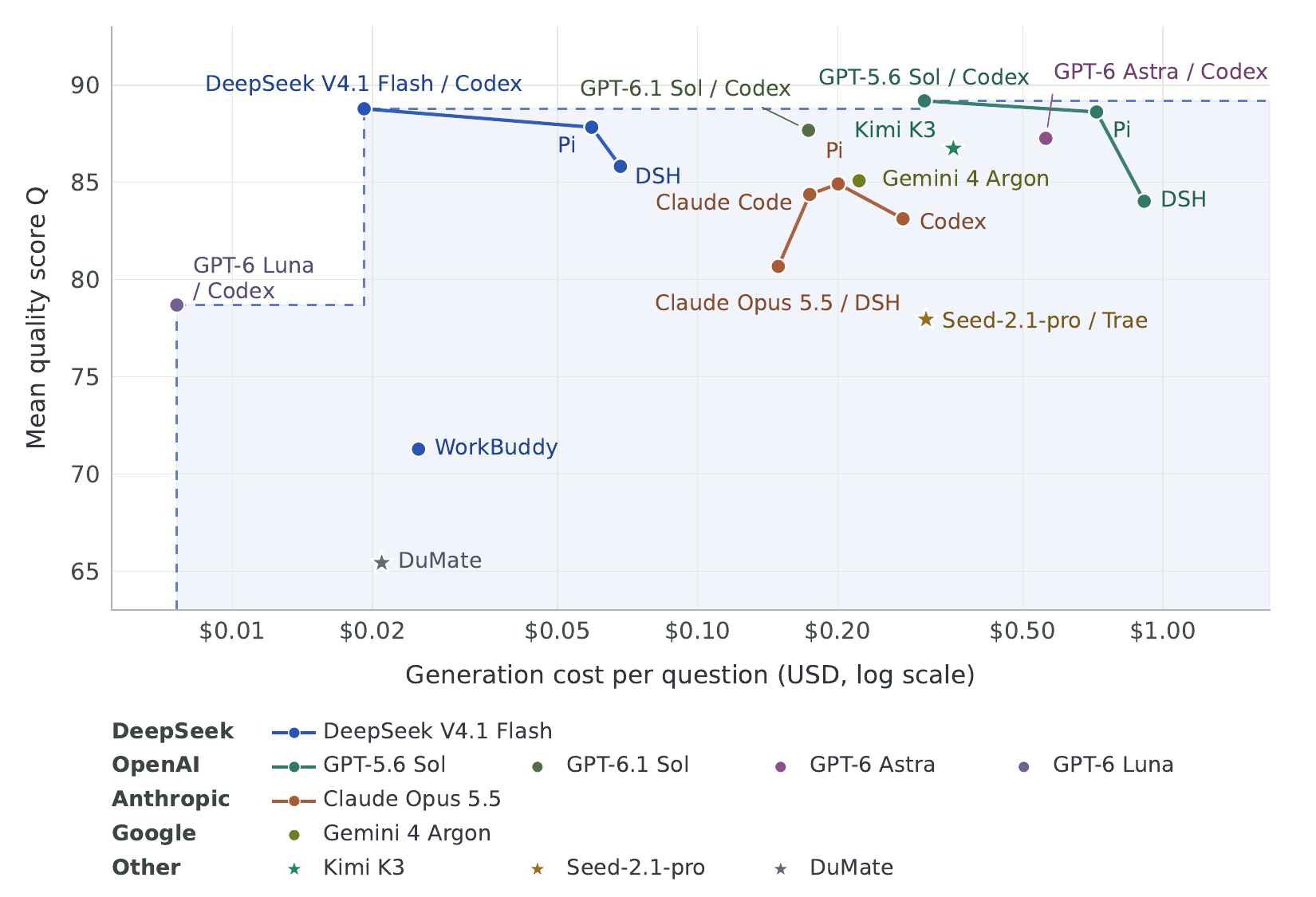}
  \caption{Generation cost per question versus mean quality score $Q$ for 18 configurations on the same 230 questions. Colors identify models; solid lines connect harness configurations of the same model. The dashed staircase and shading show the best observed quality available at each cost budget.}
\label{fig:cost-quality-11}
  \end{figure}
\endgroup

\begingroup
\renewcommand{\paratitle}[1]{\noindent\textbf{#1}\quad}
\setlength{\textfloatsep}{10pt plus 2pt minus 2pt}
\setlength{\floatsep}{8pt plus 2pt minus 1pt}
\setlength{\intextsep}{8pt plus 2pt minus 1pt}
\makeatletter
\setlength{\@fptop}{0pt}
\setlength{\@fpsep}{12pt}
\setlength{\@fpbot}{0pt plus 1fil}
\makeatother
\setlength{\abovedisplayskip}{8pt plus 2pt minus 2pt}
\setlength{\belowdisplayskip}{8pt plus 2pt minus 2pt}
\titlespacing*{\section}{0pt}{5mm plus 1mm minus 1mm}{2mm}
\titlespacing*{\subsection}{0pt}{3.5mm plus 1mm minus 0.5mm}{1.5mm}
\titlespacing*{\subsubsection}{0pt}{2.5mm}{1.2mm}

\section{Introduction}
\label{sec:intro}

Finance has become one of the most active application areas of AI agents \citep{fintrace,findeepresearch,bigfinancebench}. Requests to such agents range in type from company and industry research to macroeconomic and policy analysis, and in difficulty from numerical questions over financial reports \citep{finqa,tatqa,financebench} to in-depth research and investment planning \citep{findeepresearch,icbcbench}. Many of these requests are open-ended: they admit several defensible conclusions and have no ground-truth answer. These requests can influence real-world financial decisions, making hallucinated facts or figures potentially costly \citep{phantom,finground}.

\textit{However, existing benchmarks do not adequately capture this breadth, complexity, and open-endedness. Three challenges remain.}
\textbf{(1) Diverse task types and difficulty.}
Benchmarks focused on numerical reasoning \citep{finqa,tatqa} or financial search \citep{finsearchcomp} cover only part of real-world financial requests.
Broader coverage is needed across task types and complexity levels.
\textbf{(2) No ground-truth answer.} Many requests call for assumptions, scenarios, or trade-offs, so responses cannot be scored against a reference answer. 
\textbf{(3) Grounding versus correctness.} Rubrics do not verify individual facts \citep{finlfqa}, and whether a claim has evidence and whether it is true are separate questions. Existing methods check only one of them, either against the agent's trace \citep{deephallubench} or against external sources \citep{factscore,longfact,zhou2026cascaded}.


To address these challenges, we introduce \textbf{\bench}, a benchmark of 230 financial tasks drawn from real-world requests and spanning \textbf{8 categories, 2 languages, and 4 complexity levels}. We also propose a reference-free evaluation framework that mirrors how financial work is assessed in practice: an analysis is judged both on whether it soundly answers the request and on whether its facts hold up. Question-specific rubrics with positive and negative criteria score task completion and analytical quality. A two-stage claim audit examines the facts: claims are assessed using trace evidence, with additional verification when the available evidence is insufficient or conflicting. Audit results feed into the quality score and are also reported as separate hallucination and grounding metrics. We evaluate 18 agent configurations that pair different foundation models with different harnesses (\cref{fig:cost-quality-11}).

Our contributions are as follows:
\begin{itemize}
\item \textbf{Benchmark.} We release \bench, 230 real-world financial tasks across 8 categories, 2 languages, and 4 complexity levels, together with question-specific rubrics, the responses and execution traces of 18 agent configurations, and per-claim audit judgments.

\item \textbf{Evidence-driven evaluation framework.} We propose a reference-free framework that combines question-specific rubrics with a two-stage claim audit for hallucinations, integrates audit evidence into quality scoring, and validates the judgments against expert annotations.

\item \textbf{Empirical findings.} Across 18 configurations, we find that
(1) \textit{rubric-only scoring can overstate performance}: auditing lowers the baseline leader's score by 7.58 points and moves it to fourth place, showing that task fulfillment can mask factual errors;
(2) \textit{harness choice substantially shapes evidence grounding}: in the complete $3\times3$ comparison, harness differences account for 28.6\% of question-centered variation in verified-but-ungrounded claim rates, versus 8.8\% for model differences, underscoring the need to evaluate complete model--harness configurations; and
(3) \textit{complex tasks expose remaining capability gaps}: Quality scores fall from 89.44 on direct retrieval to 79.18 on the most complex tasks. Case studies show that agents struggle to explain how events affect company profits or stock prices, compare financial data on a consistent basis, and justify their conclusions.
\end{itemize}


\section{Related Work}
\label{sec:related}

\paratitle{Financial benchmarks.}FinQA \citep{finqa}, ConvFinQA \citep{convfinqa}, TAT-QA \citep{tatqa}, and FinanceBench \citep{financebench} test numerical reasoning and QA over financial reports and tables, and FinBen unifies financial NLP datasets \citep{finben}. Recent benchmarks target harder tasks: financial search in FinSearchComp \citep{finsearchcomp}, analyst work in Finance Agent Benchmark and FinanceQA \citep{financeagent,financeqa}, and expert-level analysis of Japanese annual reports in EDINET-Bench \citep{edinetbench}. BizFinBench and BizFinBench.v2 derive tasks from financial business data \citep{bizfinbench,bizfinbenchv2}, and the general-domain WildBench and Arena-Hard use real conversation logs \citep{wildbench,arenahard}. \bench also uses real-world requests but has no reference answers (\cref{tab:comparison}).

\begin{table*}[!htbp]
\centering
\caption{Comparison with representative benchmarks. \cmark: yes; \xmark: no; \pmark: varies by track; NR: not reported. \emph{Reference-free}: scoring needs no reference answer. \emph{Claim audit}: extracted factual statements, not predefined rubric items, are evaluated. \emph{Own trace audit}: claims are checked against the evaluated system's own execution trace. \emph{New external evidence}: the evaluator retrieves evidence beyond that trace to verify claims.}
\label{tab:comparison}
\begingroup
\fontsize{8}{10}\selectfont
\setlength{\tabcolsep}{2pt}
\renewcommand{\arraystretch}{1}
\begin{tabularx}{\textwidth}{@{}
>{\hsize=1.45\hsize\linewidth=\hsize\raggedright\arraybackslash}X
>{\hsize=1.4\hsize\linewidth=\hsize\raggedright\arraybackslash}X
>{\hsize=1.15\hsize\linewidth=\hsize\raggedright\arraybackslash}X
>{\hsize=0.75\hsize\linewidth=\hsize\centering\arraybackslash}X
>{\hsize=0.65\hsize\linewidth=\hsize\centering\arraybackslash}X
>{\hsize=0.7\hsize\linewidth=\hsize\centering\arraybackslash}X
>{\hsize=0.9\hsize\linewidth=\hsize\centering\arraybackslash}X
>{\hsize=1\hsize\linewidth=\hsize\centering\arraybackslash}X@{}}
\toprule
\textbf{Benchmark} & \textbf{Reference} & \textbf{Question source} & \textbf{Reference-free} & \textbf{Claim audit} & \textbf{Own trace audit} & \textbf{New external evidence} & \textbf{Task-specific rubric} \\
\midrule
FinSearchComp & \citep{finsearchcomp}
& Experts & \xmark & \xmark & \xmark & \xmark & Answer rules \\
\addlinespace[3pt]
BizFinBench.v2 & \citep{bizfinbenchv2}
& User requests & \pmark & \xmark & \xmark & \xmark & NR \\
\addlinespace[3pt]
PRBench (Finance) & \citep{prbench}
& Experts & \cmark & \xmark & \xmark & \xmark & \cmark \\
\addlinespace[3pt]
FinFIRST & \citep{finfirst}
& Demand-led experts & \xmark & \xmark & \xmark & \xmark & \cmark \\
\addlinespace[3pt]
FinTrace & \citep{fintrace}
& Mixed sources & \xmark & \xmark & \cmark & \xmark & Shared rubric \\
\addlinespace[3pt]
DeepResearch Bench & \citep{deepresearchbench}
& Demand-led experts & \pmark & \cmark & \xmark & \xmark & \cmark \\
\addlinespace[3pt]
ICBCBench & \citep{icbcbench}
& Industry experts & \pmark & \cmark & \xmark & \xmark & \cmark \\
\addlinespace[3pt]
DeepHalluBench & \citep{deephallubench}
& Benchmarks + adversarial tasks & \pmark & \cmark & \cmark & \xmark & Intent checks \\
\midrule
\rowcolor{fpanel}
\textbf{\textsc{AgenticFinBench}(ours)} &  
& \textbf{Production logs} & \cmark & \cmark & \cmark & \cmark & \cmark \\
\bottomrule
\end{tabularx}
\endgroup
\end{table*}

\paratitle{Long-form factuality and hallucination detection.}FActScore verifies atomic facts in long responses \citep{factscore}, SAFE does so with a search agent \citep{longfact}, and VeriScore focuses on verifiable claims \citep{veriscore}. In finance, PHANTOM tests hallucination detection in long-context QA \citep{phantom}, FinGround verifies atomic claims and corrects attribution in document QA \citep{finground}, and FinLFQA evaluates long-form answers and source attribution \citep{finlfqa}. DeepHalluBench audits actions, claims, and sub-queries in research trajectories to trace how hallucinations arise and propagate \citep{deephallubench}. \bench likewise audits claims against the system's own trace. It also checks unsupported claims against external evidence, separating true but ungrounded claims from factual errors, and feeds the results into rubric scores.

\paratitle{Rubric-based evaluation and deep research evaluation.}HealthBench, PRBench, and ResearchRubrics score open-ended responses with question-specific rubrics \citep{healthbench,prbench,researchrubrics}, and OpenRubrics derives rubrics from preference data \citep{openrubrics}. DeepResearch Bench combines adaptive criteria with citation verification \citep{deepresearchbench}, and Mind2Web 2 checks answers and sources with an agent judge \citep{mind2web2}. FinDeepResearch, BigFinanceBench, and ICBCBench apply fine-grained criteria to financial research \citep{findeepresearch,bigfinancebench,icbcbench,zhou2026activation}. For process evaluation, FinTrace scores tool calls, intermediate steps, and outputs against expert-annotated trajectories \citep{fintrace}, and FinFIRST applies atomic rubrics to information acquisition, source verification, and answer formation \citep{finfirst}. In \bench, claim-level audit evidence enters rubric scoring, so factual errors directly lower the quality score.


\section{\bench}
\label{sec:bench}

\subsection{Overview and Task Taxonomy}
\label{sec:overview}
\label{sec:dataset}
\paratitle{Task definition.}Each task in \bench is a single financial question taken from a real-world request and has no reference answer. A system must deliver an open-ended analysis and may use tools for retrieval and calculation.

\paratitle{Task taxonomy.}We classify questions by deliverable and complexity (Appendix~\ref{app:taxonomy}, \cref{tab:taxonomy}). The eight categories are knowledge Q\&A, data and calculation, screening, comparison, investment research, financial analysis, event analysis, and investment planning. Categories follow deliverables, not subjects or formats: writing a research report is investment research; retrieving one is data and calculation. Complexity (L1--L4) reflects required synthesis and decision autonomy, from direct retrieval to comprehensive research or decision design. \Cref{fig:examples} illustrates each category with a question and one positive and one negative rubric item.

\definecolor{exPos}{HTML}{568F5A}
\definecolor{exNeg}{HTML}{BF5F5B}
\tcbset{examplecard/.style={enhanced, colback=white, boxrule=0.6pt, arc=2pt,
  boxsep=1pt, left=3pt, right=3pt, top=1pt, bottom=1pt,
  toptitle=0.5pt, bottomtitle=0.5pt, before skip=0pt, after skip=2pt,
  coltitle=nnText, fonttitle=\sffamily\footnotesize,
  fontupper=\sffamily\fontsize{8}{8.5}\selectfont\raggedright\setlength{\parskip}{0pt}}}
\begin{figure}[!ht]
\centering
\input{tables/fig_examples_cards.tex}
\caption{Representative questions, one per category, with complexity level and language. Chinese questions are shown in English translation; ZH/EN denotes the original language. Each card shows one positive (+) and one negative ($-$) item from the question's rubric, translated and abridged; italics give the positive item's dimension or the negative item's severity and penalty.}
\label{fig:examples}
\end{figure}

\begin{figure}[h]
  \centering
  \includegraphics[width=\textwidth]{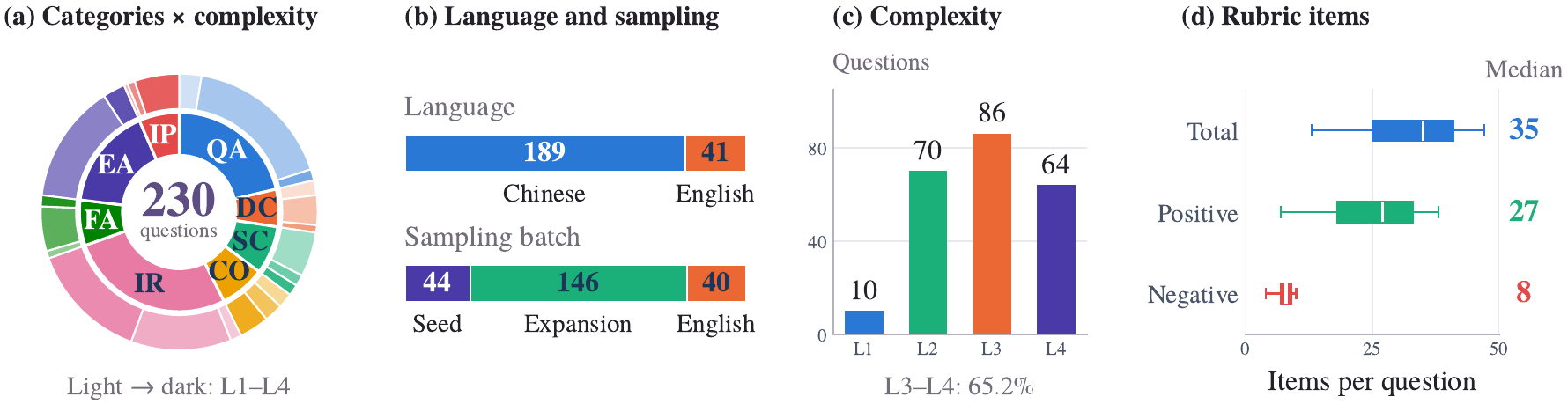}
  \caption{Dataset overview of \bench. (a) Task categories (inner ring) and complexity within each category (outer ring; light to dark denotes L1--L4); categories follow Table~\ref{tab:taxonomy}(a) in Appendix~\ref{app:taxonomy}. (b) Question language and sampling batch, shown separately because the English-language subset differs from the English batch. (c) Overall complexity distribution. L3--L4 account for 65.2\% of questions. (d) Total, positive, and negative rubric items per question. Boxes show interquartile ranges, white lines mark medians, and whiskers show minima and maxima.}
  \label{fig:distribution}
  \end{figure}

\paratitle{Data statistics.}\Cref{fig:distribution} summarizes the 230 questions, 189 in Chinese and 41 in English. Investment research (27.0\%) and knowledge Q\&A (21.3\%) are the largest categories, and L3 and L4 together account for 65.2\% of the questions.

\subsection{Construction Process}
\label{sec:construction}

We construct \bench to reflect the diversity of real-world financial information needs. The construction process combines question collection, filtering, anonymization, and expert review to produce an evaluation set spanning different task categories and complexity levels. \Cref{fig:construction} summarizes this process.

\begin{figure}[htbp]
  \centering
  \includegraphics[width=\textwidth]{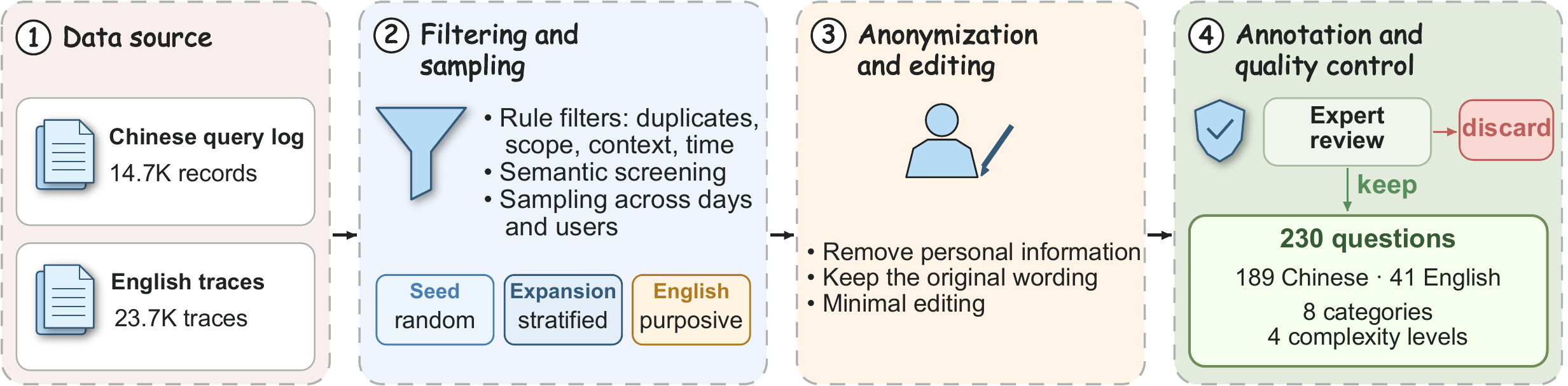}
  \caption{Construction pipeline of the evaluation set. Questions are collected, filtered, anonymized, and reviewed by experts, yielding 230 questions: 189 Chinese and 41 English, spanning 8 task categories and 4 complexity levels. }
  \label{fig:construction}
  \end{figure}

\paratitle{Data source.}The questions were curated from real-world requests. The seed and expansion batches were selected from 14,697 requests collected over a 30-day period, while the English batch was curated from a separate collection of 23,731 interaction records. Selected questions underwent\textit{ anonymization, expert review and secondary editing} before inclusion in the benchmark.

\paratitle{Filtering and sampling.}Rule-based filters remove duplicates, off-topic or context-dependent queries and requests for guaranteed returns or specific trade prices, and the remaining candidates undergo semantic screening (\cref{app:collection}). The seed batch is sampled at random across days, the expansion batch by stratified sampling with at most three questions per user, and the English batch manually for self-contained requests with checkable content. The set therefore covers real-world requests but is not a probability sample of production traffic.

\paratitle{Anonymization and editing.}Human experts remove personal user information from the selected queries and otherwise keep the original wording, including informal phrasing.

\paratitle{Annotation and quality control.}The experts then check every question by hand and remove ill-posed or unclear ones. The 230-question evaluation set contains 44, 146 and 40 questions from the seed, expansion, and English batches, respectively. 



\section{Evaluation Framework}
\label{sec:eval}

\Cref{fig:overview} shows the evaluation pipeline. Given a question $q$, a system produces a response $a$ and an execution trace $\tau$. Question-specific rubrics score the response (\cref{sec:rubric}), and a two-stage claim audit checks its factual claims (\cref{sec:audit}). Both feed a single quality score $Q$ (\cref{sec:score}). \Cref{sec:validation} validates the evaluator against human experts and a judge from another model family. All rubrics, prompts, responses, traces, and extracted claims are \emph{frozen}: they are fixed before evaluation and reused unchanged in every analysis.

\subsection{Question-Specific Rubrics}
\label{sec:rubric}
\paratitle{Rubric construction.}Following DeepResearch Bench \citep{deepresearchbench}, we generate task-specific criteria under preset dimensions and build each rubric from the question text alone. DeepSeek V4.1 Flash drafts the rubric and reviews it in a separate call, and neither call sees a system response, so no rubric is fitted to a particular response. \Cref{sec:validation} tests this overlap with a judge from another model family and also reports an expert review of the rubrics.

\paratitle{Scoring structure.}Following ResearchRubrics \citep{researchrubrics}, we split the rubric of question $q$ into positive items $\mathcal P_q$ and negative items $\mathcal N_q$. For a response $a$, binary judgments $x_{qi}(a)\in\{0,1\}$ and $z_{qj}(a)\in\{0,1\}$ indicate whether positive item $i$ is satisfied and whether negative item $j$ is triggered.
Positive items fall into four dimensions with fixed budgets $\mathbf b=(20,35,30,15)$: task completion, explanation and reasoning, evidence and accuracy, and scope and applicability. The budget $b_d$ of dimension $d$ is split equally among its items $\mathcal P_{q,d}$:
\begin{equation}
w_{qi}=\frac{b_d}{|\mathcal P_{q,d}|}\quad(i\in\mathcal P_{q,d}),
\qquad
\sum_{i\in\mathcal P_q}w_{qi}=\sum_{d=1}^{4}b_d=100.
\label{eq:rubric-weights}
\end{equation}
A satisfied positive item earns $w_{qi}$. Each negative item is assigned a severity level when the rubric is drafted, which fixes its nominal penalty $p_{qj}$ and any score cap: \emph{major} (12 points), \emph{critical} (25 points, cap 60), or \emph{fundamental} (80 points, cap 20). The fundamental level is reserved for errors that invert the question's core concept or task. The review call checks each assignment.

\begin{figure}[t]
  \centering
  \includegraphics[width=\textwidth]{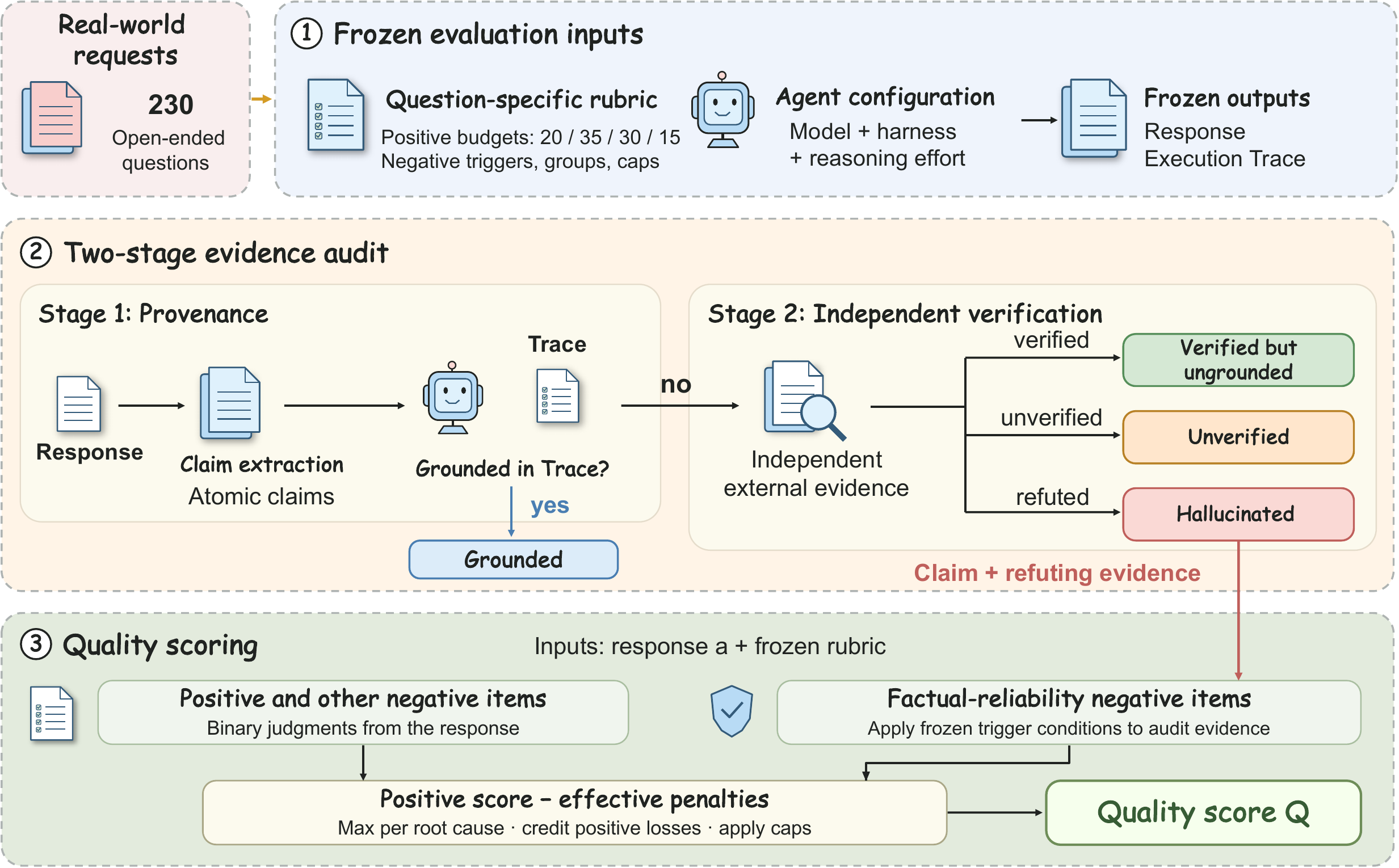}
  \caption{Overview of the \bench evaluation framework. (1) Question-specific rubrics and system responses with execution traces are frozen before evaluation. (2) Atomic claims are assessed using trace evidence, with additional verification when the available evidence is insufficient or conflicting. (3) Refuted claims trigger factual-reliability items that extend the baseline rubric; all items are combined into one quality score in which each error is charged once and critical errors cap the score.}
  \label{fig:overview}
  \end{figure}

\subsection{Two-Stage Claim Audit}
\label{sec:audit}
\paratitle{Audit design.}The audit separates grounding from factual correctness: each claim is assessed using trace evidence, with additional verification when the available evidence is insufficient or conflicting. Before the audit, we extract a set of atomic claims $C(a)=\{c_1,\dots,c_m\}$ from response $a$. Each claim can be checked independently and keeps the key qualifiers of the original text. Claims are extracted once per response (Appendix~\ref{sec:claim-extraction}).

\begin{table}[H]
  \centering
  \begin{minipage}{0.65\textwidth}
  \centering
  \caption{The four claim outcomes of the two-stage audit.}
  \label{tab:outcomes}
  \footnotesize
  \setlength{\tabcolsep}{4pt}
  \begin{tabular}{@{}lll@{}}
  \toprule
  Outcome & \makecell[l]{Stage 1:\\provenance} & \makecell[l]{Stage 2:\\verification} \\
  \midrule
  \textcolor{cGround}{\rule{1.6ex}{1.6ex}} Grounded & grounded & \makecell[l]{verified /\\refuted /\\unverified} \\
  \textcolor{cVU}{\rule{1.6ex}{1.6ex}} Verified but ungrounded & ungrounded & verified \\
  \textcolor{cUnv}{\rule{1.6ex}{1.6ex}} Unverified & ungrounded & unverified \\
  \textcolor{cHal}{\rule{1.6ex}{1.6ex}} Hallucinated & ungrounded & refuted \\
  \bottomrule
  \end{tabular}
  \par\smallskip
  {\raggedright\footnotesize Grounded claims also receive a factual verdict, which may reuse trace evidence; their diagnostic category remains grounded regardless of that verdict.\par}
  \end{minipage}
  \end{table}

\paratitle{Stage 1: provenance.}The auditor checks whether the complete trace supports all substantive content and key qualifiers of a claim, giving $P(c,\tau)\in\{\textit{grounded},\textit{ungrounded}\}$. Valid evidence includes tool and API outputs, retrieved text, and calculations whose operands appear in the trace. The system's own reasoning, plans, or restatements do not count.

\paratitle{Stage 2: verification.}The auditor reuses reliable trace material and seeks additional sources or calculations when the available evidence is insufficient or conflicting, giving $V(c)\in\{\textit{verified},\textit{refuted},\textit{unverified}\}$. Official documents and primary data take priority. A claim is refuted only when the evidence explicitly conflicts with it; without sufficient evidence, it remains unverified. Relative expressions such as ``latest'' refer to each response's generation time, not the original query time. Each claim keeps its stated period, and the auditor refutes a claim only with evidence for that same period. The two stages yield the four outcomes in \cref{tab:outcomes}; Appendix~\ref{sec:audit-agent} gives the audit instructions and output fields.

\paratitle{Hallucination criterion.}A claim is hallucinated only if it is ungrounded and externally refuted, i.e., $P(c,\tau)=\textit{ungrounded}$ and $V(c)=\textit{refuted}$; ungrounded or unverified claims alone are not hallucinations. Conversely, grounded means backed by the trace, not correct: if a response repeats an erroneous tool output, the inherited error is still grounded, as it originates in the source rather than the model (\cref{sec:cases}). The auditor also records a factual verdict for grounded claims, which may reuse trace material. Grounded claims retain their diagnostic category, but confirmed factual errors may still affect the quality score. Both stages are needed: if being ungrounded alone counted as hallucination, 42.72\%--91.45\% of the flagged claims would be false alarms, since they turn out to be verified (Appendix~\ref{sec:rq-stages}).

\subsection{Quality Score and Diagnostic Metrics}
\label{sec:score}
\paratitle{Factual-reliability penalties.}Each refuted claim, grounded or not, is rechecked against its saved refuting evidence and the full response. If this check finds that the evidence does not contradict the assertion actually made in the response, the refutation is withdrawn and cannot trigger a factual-reliability penalty. For affected ungrounded claims, the diagnostic outcome is conservatively changed to unverified, rather than automatically to verified. Original audit labels are retained only for provenance. Each confirmed factual error is linked to a baseline negative item or, if not already fully covered, to an extension item whose severity depends on the error's effect on the response. Claims sharing the same error root are penalized once. Extension items use the major and critical levels of \cref{sec:rubric} plus a lower \emph{minor} level. A minor error (3 points) does not weaken what the question asks for. A major error (12 points) weakens a substantive fact, explanation, or comparison. A critical error (25 points, cap 60) reverses the core conclusion, invalidates necessary evidence, or breaks a key calculation.  Only refuted claims trigger these penalties, and an error already fully covered by a baseline negative item is penalized only through that item. The audit also rechecks positive items: a positive item becomes unsatisfied only when an error invalidates its entire pass condition.

\paratitle{Quality score.}Penalties that stem from the same error form one root-cause group and are charged once. The group's penalty is the highest nominal penalty $p_{qj}$ among its triggered items ($z_{qj}=1$), minus any positive weight the error has already removed. Let $W_q=\sum_{i\in\mathcal P_q}w_{qi}x_{qi}$ be the positive score after the audit recheck. The quality score of question $q$ is
\begin{equation}
Q_q=\min\!\left\{M_q,\;\max\!\left[0,\;W_q-D_q^{\mathrm{rub}}-D_q^{\mathrm{hall}}\right]\right\},
\end{equation}
where $D_q^{\mathrm{rub}}$ and $D_q^{\mathrm{hall}}$ are the effective penalties of the baseline and factual-reliability groups, and $M_q$ is the strictest cap among triggered items (100 if none applies). The baseline score $B_q$ applies the same aggregation to the rubric-only judgments and omits the audit recheck and factual-reliability items; $Q-B$ therefore isolates the effect of the audit. Appendix~\ref{app:rubric-example} gives the full calculation and \cref{fig:case-q} a worked example. The score of configuration $s$ is $Q(s)=\frac{1}{N}\sum_{q\in\mathcal T}Q_{q,s}$, where $\mathcal T$ is the benchmark question set and $N=230$. Confidence intervals come from a question-level bootstrap.

\paratitle{Diagnostic metrics.}Diagnostic metrics use the same reconciled claim verdicts that determine eligibility for factual-reliability penalties, and are reported separately from $Q$. At the response level, we report the share of a configuration's 230 responses that contain at least one hallucinated claim (responses with hallucination). At the claim level, the four outcomes partition the claims of a configuration, and their shares are the grounded rate, the verified-but-ungrounded (VU) rate, the unverified rate, and the hallucination rate, which sum to 100\%. The unconfirmed rate $U$ is the sum of the hallucination and unverified rates and is an upper bound on the hallucination rate as defined here. These rates pool all claims of a configuration, whereas paired comparisons between configurations use per-question shares averaged with equal weight.

\subsection{Validating the Evaluator}
\label{sec:validation}
\paratitle{Agreement with human experts.}We sampled 1800 responses (100 questions $\times$ 18 configurations) uniformly at random, and from them 2,126 claims for Stage~1, 1,372 ungrounded claims for Stage~2, and 2,955 baseline rubric items. Two of three senior financial experts labeled each sample independently and blind to model identity, and the third resolved disagreements. \Cref{tab:annotation-protocol} reports raw agreement and Cohen's $\kappa$ with the adjudicated labels and between experts. Across all four validation tasks, evaluator agreement with adjudicated expert labels is comparable to inter-expert agreement in both raw agreement and  Cohen's $\kappa$.

\paratitle{Evaluator overlap.}DeepSeek V4.1 Flash judges the rubric items and also underlies four evaluated configurations (Codex CLI, Pi, DSH, and WorkBuddy), so its judgments may favor their responses. We therefore re-judged the baseline rubric items of 100 questions for the original 18 configurations with MiMo-V2.6, a judge from a model family outside the evaluated configurations, using the frozen rubrics and the same aggregation. Measured as a difference-in-differences of baseline scores between the two judges and between the DeepSeek and other configurations, the self-preference effect is $\delta=0.11$ points (95\% CI $[-0.87, 0.63]$), and the Kendall rank correlation between the two judges' configuration rankings is 0.89. 

\begin{table}[!htbp]
  \def\pairedvalidation{1}
  \begin{minipage}[t]{0.49\textwidth}
  \vspace{0pt}
  \ifdefined\pairedvalidation\else\begin{table}[!htbp]\fi
\centering
\caption{Agreement with adjudicated expert labels and between experts. Rates are agreed/reviewed (\%).}
\label{tab:annotation-protocol}
\ifdefined\pairedvalidation\fontsize{7.2}{9}\selectfont\setlength{\tabcolsep}{1.3pt}\else\small\fi
\renewcommand{\arraystretch}{1.1}
\begin{tabularx}{\linewidth}{@{}Xrrrrrrr@{}}
\toprule
 & & \multicolumn{3}{c}{\makecell{Evaluator vs. experts}} & \multicolumn{3}{c}{\makecell{Between experts}} \\
\cmidrule(lr){3-5}\cmidrule(l){6-8}
Task & $n$ & Agreed & \% & $\kappa$ & Agreed & \% & $\kappa$ \\
\midrule
\makecell[l]{Stage 1: provenance} & 2,126 & 1,963 & 92.33 & 0.81 &1,980 & 93.13 & 0.86 \\
\makecell[l]{Stage 2: verification} & 1,372 & 1,257 & 91.62 & 0.82 & 1,239& 90.31 & 0.81 \\
\makecell[l]{Positive rubric} & 1,827 & 1,592 & 87.14 & 0.73 & 1,578& 86.37 &  0.73\\
\makecell[l]{Negative rubric} & 1,128 & 957 & 84.84 & 0.70 & 963& 85.37 &  0.71 \\
\bottomrule
\end{tabularx}
\ifdefined\pairedvalidation\else\end{table}\fi

  \end{minipage}\hfill
  \begin{minipage}[t]{0.49\textwidth}
  \vspace{0pt}
  
\ifdefined\pairedvalidation\else\begin{table}[!htbp]\fi
\centering
\caption{Agreement with adjudicated expert preferences. Rates are agreed/reviewed pairs (\%).}
\label{tab:expert-preferences}
\ifdefined\pairedvalidation\fontsize{7.8}{10}\selectfont\setlength{\tabcolsep}{2pt}\else\small\fi
\renewcommand{\arraystretch}{1.1}
\begin{tabular}{@{}lcc@{}}
\toprule
Scoring method & \makecell{Random\\pairs} & \makecell{Ranking-reversal\\pairs} \\
\midrule
Baseline $B$ & 67.8 & 19.9\\
Simple penalty & 71.3 & 54.7 \\
Final $Q$ & 73.3 & 66.1 \\
\bottomrule
\end{tabular}

\ifdefined\pairedvalidation\else\end{table}\fi

  \end{minipage}
  \end{table}

\paratitle{Expert preference alignment.}Two of three experts independently compared each answer pair, blinded to model identities, automated scores, and audit judgments, and the third resolved disagreements; ties were allowed, and a method agrees only if it gives the same verdict. On random pairs (\cref{tab:expert-preferences}), $Q$ agrees with experts on 73.3\%, 5.5 points above $B$ (95\% CI $[2.8, 8.2]$) and close to the 71\% agreement between experts. On ranking-reversal pairs, which $B$ and $Q$ order oppositely,  experts side with $Q$ on 66.1\% and with $B$ on 19.9\%, and $Q$ exceeds a simple penalty of 8 points per hallucinated claim by 11.4 points (95\% CI $[4.1, 18.7]$). When the audit changes the order, experts thus usually side with it, and its graded penalties outperform a simple penalty.


\section{Experimental Results and Analysis}
\label{sec:experiments}
\label{sec:analysis}


\subsection{Experimental Setup}
\label{sec:setup}

\paratitle{Agent configurations.}The unit of comparison is the \textbf{agent configuration}: a foundation model, a harness, and a reasoning effort (\cref{tab:configs} in Appendix~\ref{app:generation}). The harness is the program that drives the model; it determines which tools are available, whether the model retrieves, and what enters the context and the trace. The 18 configurations include comparisons of six models (GPT-6 Astra, GPT-5.6 Sol, GPT-6 Luna, GPT-6.1 Sol, DeepSeek V4.1 Flash, and Claude Opus 5.5) on Codex CLI; DeepSeek V4.1 Flash on four harnesses (Codex CLI, Pi, DSH, and WorkBuddy); GPT-5.6 Sol on three harnesses (Codex CLI, Pi, and DSH); and Claude Opus 5.5 on four harnesses (Claude Code CLI, Codex CLI, Pi, and DSH). The other four are Gemini 4 Argon on Gemini CLI, Kimi K3 on the Kimi harness, DuMate on DuMate CLI, and Seed-2.1-pro on Trae Work.

\paratitle{Data and evaluation implementation.}The evaluation uses 4,140 frozen responses from the 18 configurations on the 230 questions. Claims are extracted by DeepSeek V4.1 Flash and audited by DeepSeek V4 Flash on DSH 0.1.5-rc.1 with high reasoning effort and subagents disabled. The Kimi K3 audit uses a budget of 1,800 seconds per question with up to three retries. Baseline rubric items and extension items are judged by DeepSeek V4.1 Flash. The final scores reuse the frozen responses, item judgments, and audit evidence, and every per-question score is checked by an independent recomputation. All responses are audited and scored. Quality scores and the change from $B$ to $Q$ are averaged over the 230 questions with equal weight; 95\% intervals come from 10,000 paired question-level bootstrap resamples. In \cref{fig:cost-quality-11}, fourteen configurations use the mean recorded-token cost of these 230 questions; Kimi, Trae, and DuMate retain their original run-average cost coordinates, and WorkBuddy retains the original visible-text cost estimate. Supplementary analyses are given in Appendix~\ref{app:rq-supplement}.

\subsection{Main Results}
\label{sec:main-results}

\Cref{tab:main} highlights four findings.
(1) \textit{Quality varies widely, but the leading scores are close.} Mean $Q$ ranges from 65.42 to 89.20, with GPT-5.6 Sol on Codex CLI highest and the top three configurations within 0.6 points.
(2) \textit{The same model performs differently across harness configurations.} DeepSeek V4.1 Flash spans 71.27--88.79 in mean $Q$ across four harnesses, a 17.52-point gap.
(3) \textit{Near-leading quality is achievable with shorter generation times.} Sol on Pi averages 2.06 minutes per response versus 6.25 on Codex, with only a 0.57-point difference in mean $Q$.
(4) \textit{Costs vary substantially without following the quality ranking.} Estimated token costs span roughly 120-fold (\$0.0076--\$0.9125 per response); the most expensive configuration, Sol on DSH, ranks twelfth in quality.

\begin{table}[H]
\centering
\caption{Quality and generation efficiency for 18 configurations on the same 230 questions. $Q$ is the quality score. Intervals are 95\% paired question-bootstrap intervals from 10,000 resamples. Rows follow mean $Q$. Bold marks the best result and underlining the second and third best in $Q$ (higher is better), time, and cost (lower is better).}
\label{tab:main}
\small
\renewcommand{\arraystretch}{0.8}
\setlength{\tabcolsep}{1pt}
\begin{tabular*}{\textwidth}{@{\extracolsep{\fill}}rllrrrr@{}}
\toprule
Rank & Model & Harness & {\boldmath$Q$} & 95\% CI & \makecell{Time\\(min)} & \makecell{Cost\\(USD)} \\
\midrule
1 & GPT-5.6 Sol & Codex CLI & \textbf{89.20} & 87.61--90.71 & 6.25 & 0.3073 \\
2 & DeepSeek V4.1 Flash & Codex CLI & \underline{88.79} & 87.44--90.09 & 14.71 & \underline{0.0192} \\
3 & GPT-5.6 Sol & Pi & \underline{88.62} & 87.31--89.89 & 2.06 & 0.7207 \\
4 & DeepSeek V4.1 Flash & Pi & 87.84 & 85.87--89.64 & 2.20 & 0.0592 \\
5 & GPT-6.1 Sol & Codex CLI & 87.68 & 85.98--89.31 & 3.60 & 0.1731 \\
6 & GPT-6 Astra & Codex CLI & 87.27 & 85.69--88.84 & 2.10 & 0.5604 \\
7 & Kimi K3 & Kimi hosted & 86.74 & 84.39--88.87 & 6.06 & 0.3546 \\
8 & DeepSeek V4.1 Flash & DSH & 85.82 & 83.67--87.87 & 3.20 & 0.0682 \\
9 & Gemini 4 Argon & Gemini CLI & 85.09 & 83.32--86.79 & 3.89 & 0.2224 \\
10 & Claude Opus 5.5 & Pi & 84.92 & 83.47--86.39 & \textbf{0.94} & 0.2005 \\
11 & Claude Opus 5.5 & Claude Code CLI & 84.38 & 82.56--86.17 & \underline{1.31} & 0.1741 \\
12 & GPT-5.6 Sol & DSH & 84.03 & 82.11--85.84 & 3.27 & 0.9125 \\
13 & Claude Opus 5.5 & Codex CLI & 83.13 & 81.40--84.80 & \underline{1.32} & 0.2764 \\
14 & Claude Opus 5.5 & DSH & 80.68 & 78.77--82.50 & 1.34 & 0.1491 \\
15 & GPT-6 Luna & Codex CLI & 78.69 & 76.74--80.67 & 2.04 & \textbf{0.0076} \\
16 & Seed-2.1-pro & Trae Work & 77.95 & 74.69--80.99 & 10.50 & 0.3098 \\
17 & DeepSeek V4.1 Flash & WorkBuddy & 71.27 & 68.39--74.09 & 9.45 & 0.0251 \\
18 & DuMate & DuMate CLI & 65.42 & 62.43--68.32 & 8.55 & \underline{0.0209} \\

\bottomrule
\end{tabular*}
\end{table}

\subsection{Analysis}
\label{sec:rq-quality}
\label{sec:audit-score-change}

\subsubsection{RQ1: How does accounting for hallucinations affect system comparisons?}
Hallucination auditing can overturn advantages under rubric-only scoring. DeepSeek V4.1 Flash on Pi loses 7.58 points and falls from first under $B$ to fourth under $Q$, whereas GPT-5.6 Sol on Codex CLI loses 0.27 points and rises from fifth to first. GPT-6.1 Sol and GPT-6 Astra rise two and three places, respectively (\cref{tab:main,fig:fulfillment}). This contrast extends beyond individual systems: the eight configurations with hallucinated claims in at least 16.5\% of responses lose 2.54--7.58 points, while the four GPT configurations on Codex CLI stay at or below 3.1\% and lose at most 0.27 points. Rubric-only scores can thus overstate performance when task fulfillment rests on incorrect facts. Adjacent ranks need not differ significantly. Across three evaluation runs on the same 100 questions with other settings fixed, the within-configuration standard deviation of mean $Q$ averaged 0.26 points. Evaluator accuracy is assessed in \cref{sec:validation}.

\begin{figure}[!htbp]
\centering
\includegraphics[width=\textwidth]{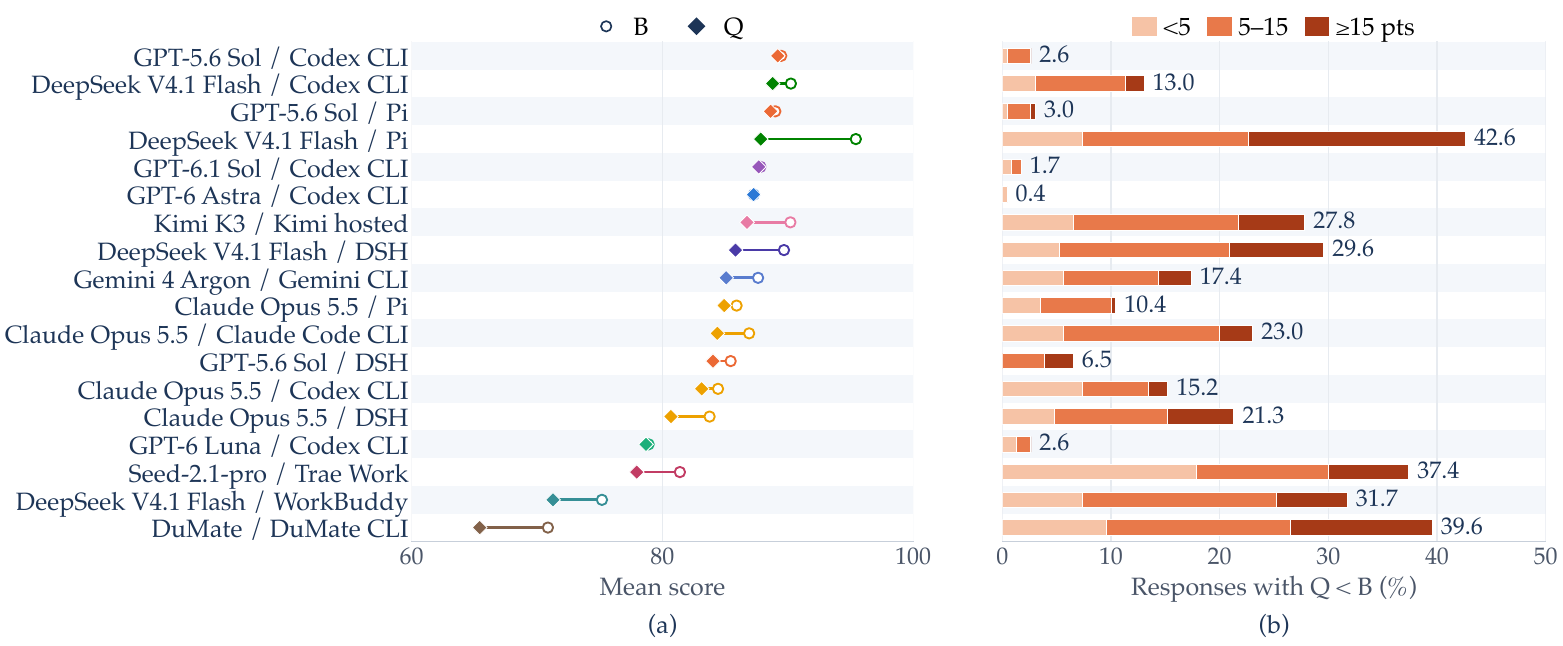}
\caption{Effect of the audit on scores across the 230 questions, rows ordered by mean $Q$. (a) Mean baseline score $B$ (circles) and quality score $Q$ (diamonds). (b) Share of responses with $Q<B$, split by deduction size in points. Scoring rules are in \cref{sec:score}.}
\label{fig:fulfillment}
\end{figure}

\begin{figure}[!htbp]
  \centering
  \includegraphics[width=\linewidth]{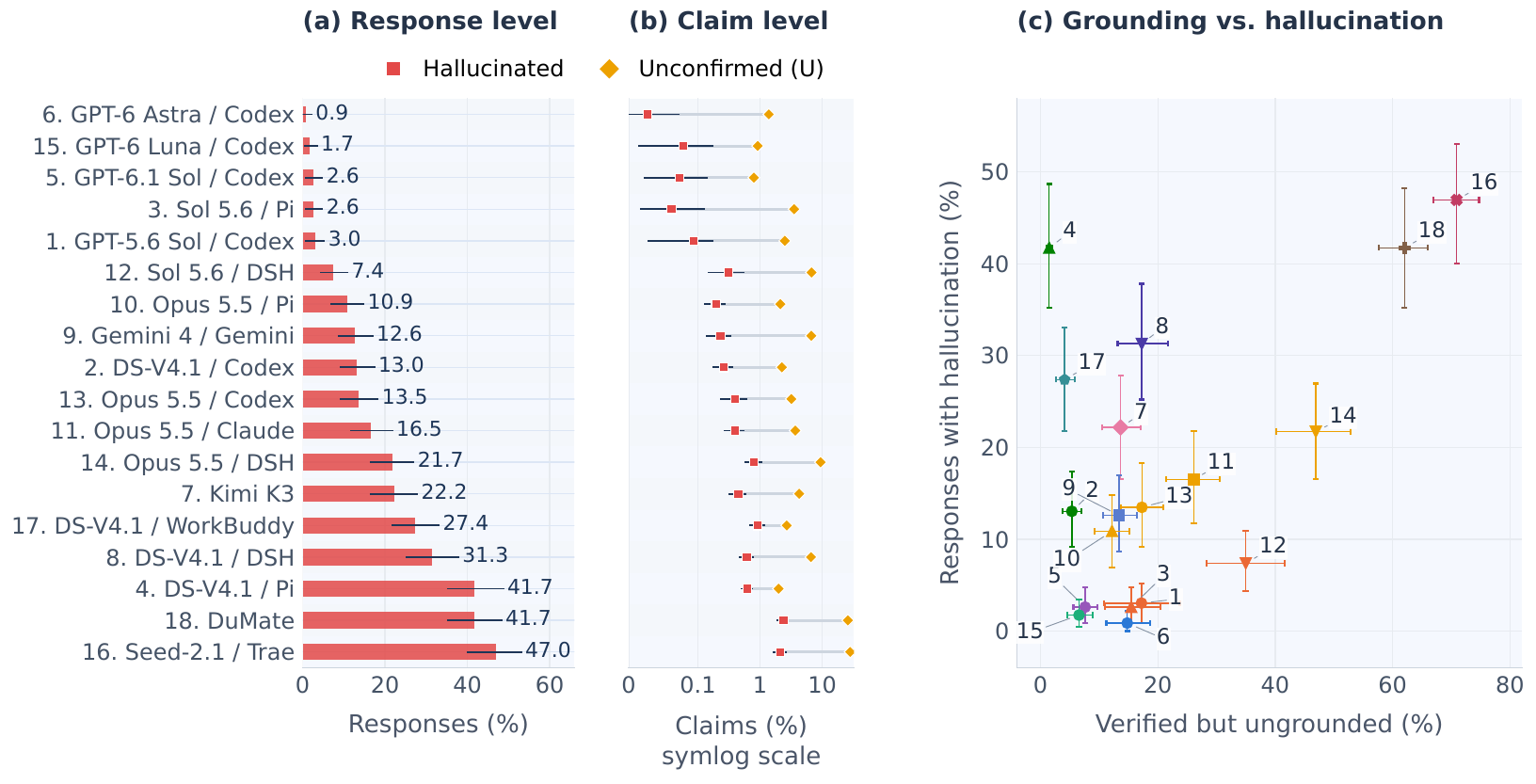}
  \caption{Audit diagnostics for 18 configurations on the 230 questions. (a) Share of responses with a hallucinated claim, with 95\% question-bootstrap intervals. (b) Claim-level hallucination rate (squares) and unconfirmed rate $U$ (diamonds). Rows in (a)--(b) are ordered by (a). (c) VU rate versus response-level hallucination rate, with 95\% intervals; numbers in (a) and (c) identify the same configurations and follow the quality-ranked rows of Table~\ref{tab:main}. Marker shapes encode harness families. Hallucinated claims lack trace support and are externally refuted; unresolved claims are not counted as hallucinations.}
  \label{fig:main}
  \end{figure}

  \subsubsection{RQ2: How prevalent are hallucinations across financial agent configurations?}
\label{sec:rq-grounding}
\label{sec:grounding-analysis}
Hallucinations occur in all 18 configurations, and low claim-level rates can mask substantial response-level exposure. Rates range from 0.03\% to 2.38\% of claims, but 0.9\%--47.0\% of responses contain at least one hallucinated claim (\cref{fig:main}). This includes 3.0\% of responses from the top-ranked GPT-5.6 Sol on Codex CLI and 41.7\% from the fourth-ranked DeepSeek V4.1 Flash on Pi. Both measures are therefore needed to capture error density and the share of responses affected. Low hallucination rates also do not imply complete trace grounding. GPT-6 Astra has hallucinations in only 0.9\% of responses, but 14.78\% of its claims are verified but ungrounded; DeepSeek on Pi shows the opposite pattern, at 41.7\% and 1.48\%, respectively (\cref{fig:main}(c)). External verification establishes support outside the trace, not that the system retrieved supporting evidence. Our hallucination measure counts only ungrounded, externally refuted claims and is therefore not a complete factual-error rate. Full statistics appear in \cref{tab:audit-full,tab:paired,tab:structure} in the appendix.

  \begin{figure}[!htbp]
    \centering
    \begin{minipage}[t]{0.52\textwidth}
    \vspace{0pt}
    \centering
\captionof{table}{Recorded tool use during response generation for the 18 configurations. Each configuration covers the 230 questions; full configurations are listed in \cref{tab:configs}. Dashes indicate partial client traces for which comparable call counts are unavailable; zero denotes no recorded call in a complete trace. Call counts are not equal units of retrieval effort.}
\label{tab:behavior}
\small
\setlength{\tabcolsep}{5pt}
\renewcommand{\arraystretch}{0.6}
\begin{adjustbox}{max width=\linewidth}
\begin{tabular}{@{}lcc@{}}
\toprule
Agent configuration & \makecell{Tool calls\\per response\\(median)} & \makecell{Responses with\\no recorded\\calls} \\
\midrule

GPT-5.6 Sol / Codex CLI & 7 & 7.0\% \\
DeepSeek V4.1 Flash / Codex CLI & 75 & 2.6\% \\
GPT-5.6 Sol / Pi & 19.5 & 15.2\% \\
DeepSeek V4.1 Flash / Pi & 20 & 0.4\% \\
GPT-6.1 Sol / Codex CLI & 5 & 0.4\% \\
GPT-6 Astra / Codex CLI & 3 & 1.3\% \\
Kimi K3 / Kimi hosted & 3 & 5.2\% \\
DeepSeek V4.1 Flash / DSH & 24 & 6.1\% \\
Gemini 4 Argon / Gemini CLI & 4 & 14.8\% \\
Claude Opus 5.5 / Pi & 3 & 5.7\% \\
Claude Opus 5.5 / Claude Code CLI & 7 & 9.6\% \\
GPT-5.6 Sol / DSH & 15 & 23.9\% \\
Claude Opus 5.5 / Codex CLI & 0 & 56.5\% \\
Claude Opus 5.5 / DSH & 1 & 33.0\% \\
GPT-6 Luna / Codex CLI & 4 & 1.3\% \\
Seed-2.1-pro / Trae Work & --- & --- \\
DeepSeek V4.1 Flash / WorkBuddy & 22 & 0.0\% \\
DuMate / DuMate CLI & --- & --- \\

\bottomrule
\end{tabular}
\end{adjustbox}

    \end{minipage}\hfill
    \begin{minipage}[t]{0.46\textwidth}
    \vspace{0pt}
    \centering
    \includegraphics[width=\linewidth]{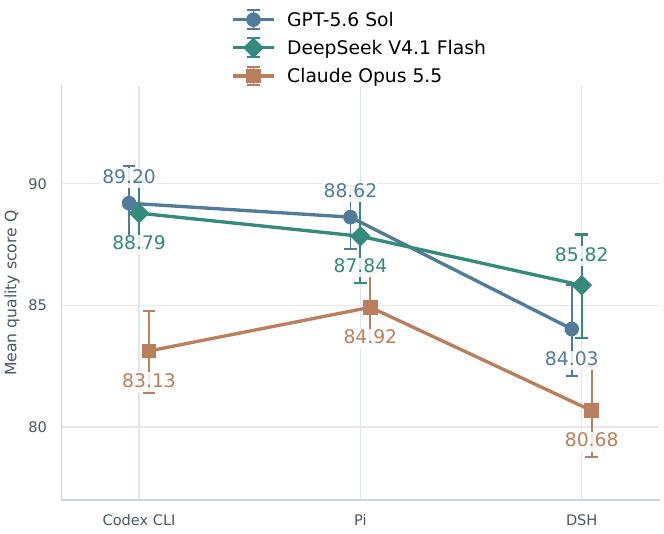}
    \caption{Model--harness comparison on the same 230 questions. Lines connect the mean quality scores of each model across Codex CLI, Pi, and DSH; bars show pointwise 95\% question-bootstrap intervals. Paired contrasts are computed from joint question resamples, not from overlap of these marginal intervals. Generation instructions, tools, and execution settings differ between harnesses.}
    \label{fig:harness-vu}
    \end{minipage}
    \end{figure}


\subsubsection{RQ3: How do foundation models and agent harnesses jointly affect performance?}
\label{sec:rq-harness}
\label{sec:rq-model}
The relative contributions of models and harnesses depend on the metric. In the complete $3\times3$ comparison of GPT-5.6 Sol, DeepSeek V4.1 Flash, and Claude Opus 5.5 on Codex CLI, Pi, and DSH, harness differences account for more of the question-centered sum of squares in per-question VU rates than model differences (28.60\% versus 8.78\%). For $Q$, the ordering reverses: the harness and model components account for 2.38\% and 3.74\%, respectively. Harness configurations also differ in quality. Averaged over the three models, Codex CLI and Pi exceed DSH by 3.53 and 3.62 points (\cref{fig:harness-vu}). Codex CLI leads for GPT and DeepSeek, while Pi leads for Opus, supporting evaluation of model--harness combinations. These comparisons cover complete configurations, including instructions, tools, and reasoning settings.

\begin{figure*}[!htbp]
  \centering
  \includegraphics[width=\textwidth]{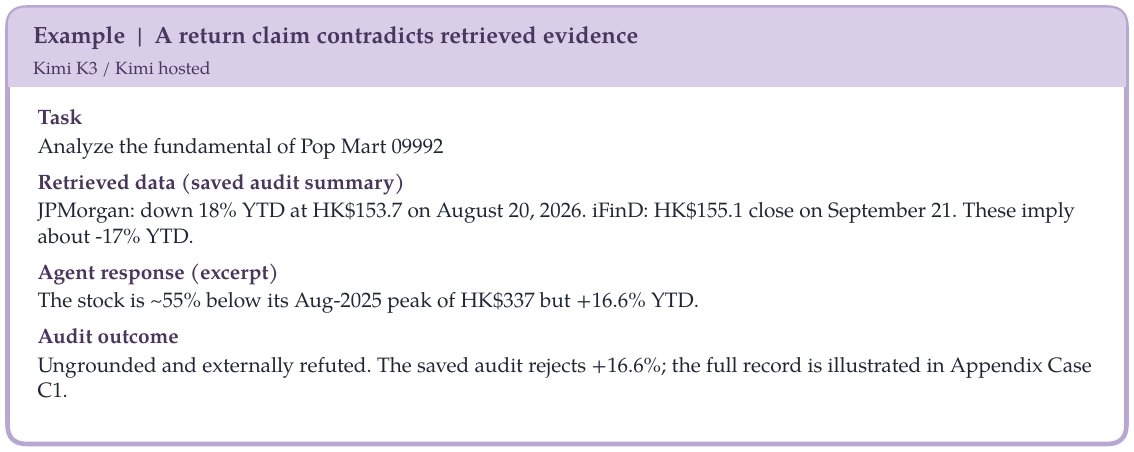}
  \caption{An example of how claim-level audit evidence informs rubric judgments and the final quality score.}
  \label{fig:inline-evidence-case}
  \end{figure*}

\begin{figure}[!htbp]
  \centering
  \begin{minipage}[t]{0.49\textwidth}
  \vspace{0pt}
  \centering
  \includegraphics[width=\linewidth]{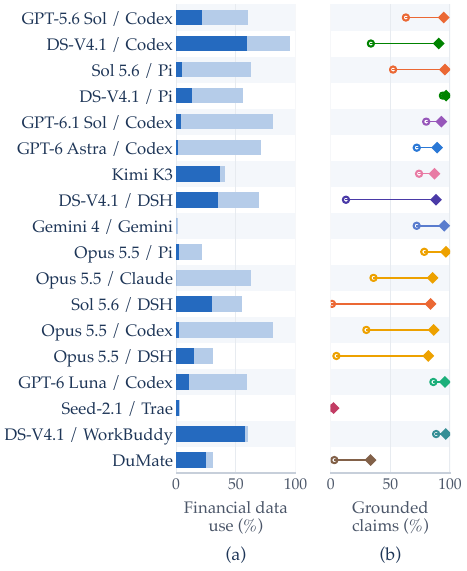}
  \caption{Financial data use for all 18 configurations on the 230 questions, rows as in \cref{fig:fulfillment}. (a) Share of responses with recorded financial-data access: tools, APIs, or code libraries (dark), plus official or financial-portal sources (light). (b) Share of grounded claims in responses without (circles) and with (diamonds) financial data, as raw means within each configuration. }
  \label{fig:tooling}
  \end{minipage}\hfill
  \begin{minipage}[t]{0.49\textwidth}
  \vspace{0pt}
  \centering
  \includegraphics[width=\linewidth]{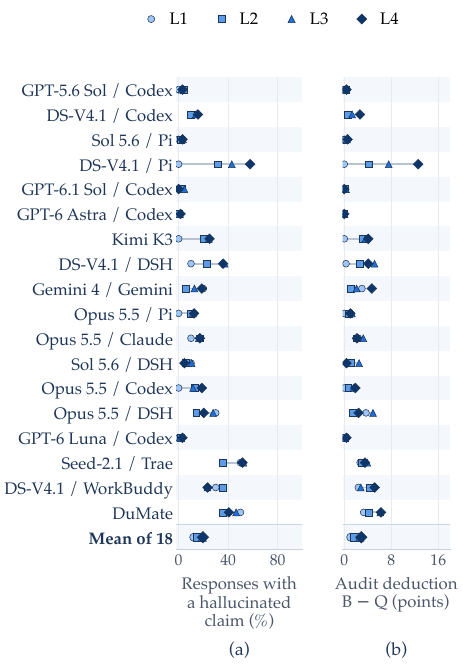}
  \caption{Factual reliability by complexity level on the 230 questions, for all 18 configurations. Rows follow Table~\ref{tab:main}. (a) Share of responses with a hallucinated claim. (b) Mean audit deduction $B-Q$. The last row averages the 18 configurations with equal weight.}
  \label{fig:difficulty}
  \end{minipage}
  \end{figure}

\subsubsection{RQ4: How is financial data use associated with evidence grounding and hallucinations?}
\label{sec:rq-tooling}
Financial data use is associated with stronger grounding, but agents can still misuse retrieved evidence. We measure recorded access through tools, APIs, code libraries, or official and financial-portal sources, separately from configured tool availability (\cref{tab:configs}). Across 18 configurations, the adjusted grounding difference is 26.3 percentage points after controlling for configuration and category, and 8.3 points among responses with recorded evidence retrieval. The raw difference is positive in all 18 configurations (\cref{fig:tooling}). \Cref{fig:inline-evidence-case} shows a response contradicting retrieved evidence (full audit in \cref{fig:cases-hal}). These associations do not establish a causal benefit of dedicated financial tools.

\subsubsection{RQ5: Where do current systems fall short across categories and complexity levels?}
\label{sec:rq-difficulty}
Complex tasks expose gaps in both answer quality and factual reliability, with distinct weaknesses across task categories. Across the 18 configurations, mean $Q$ falls from 89.44 at L1 to 79.18 at L4, the share of responses with a hallucinated claim rises from 12.2\% to 19.6\%, and mean audit deductions increase from 1.09 to 2.90 points (\cref{tab:difficulty,fig:difficulty}). Response-level hallucination rates are higher on L3--L4 than on L1--L2 for 16 of 18 configurations. By category, data and calculation has the highest mean $Q$ (90.42), investment planning the lowest (78.88), and comparison the highest response-level hallucination rate (22.5\%). These differences depend on the selected questions and rubrics, rather than a separately validated difficulty scale. An exploratory analysis of frozen reasoning criteria finds that explaining how events affect profits or prices and comparing and combining financial information account for the largest shares of unmet requirements (45.83\% and 17.07\%, totaling 62.90\%; \cref{fig:reasoning-gaps}). Quantitative modeling has the highest within-category unmet rate (27.59\%). Definitions and denominators appear in Appendix~\ref{app:reasoning-gaps}.

\begin{table}[!htbp]
\centering
\caption{Mean quality score $Q$ by category and complexity level on the 230 questions. Rows follow \cref{tab:main}; the best configuration in each column is in bold, and the last row averages the 18 configurations. Categories and levels use the agent-reviewed labels of \cref{tab:taxonomy}.}
\label{tab:difficulty}
\scriptsize
\begin{tabular}{@{}l rrrrrrrr c rrrr@{}}
\toprule
& \multicolumn{8}{c}{Category} & & \multicolumn{4}{c}{Complexity level} \\
\cmidrule(lr){2-9}\cmidrule(l){11-14}
Agent configuration & \makecell{Knowl.\\Q\&A} & \makecell{Data \&\\calc.} & \makecell{Screen-\\ing} & \makecell{Compar-\\ison} & \makecell{Invest.\\research} & \makecell{Financial\\analysis} & \makecell{Event\\analysis} & \makecell{Invest.\\planning} & & L1 & L2 & L3 & L4 \\
\midrule

\textit{Questions} & \textit{49} & \textit{14} & \textit{18} & \textit{17} & \textit{62} & \textit{17} & \textit{38} & \textit{15} & & \textit{10} & \textit{70} & \textit{86} & \textit{64} \\
\midrule
GPT-5.6 Sol / Codex CLI & 90.35 & \textbf{95.71} & 87.19 & 91.23 & 90.13 & 85.94 & 88.02 & 82.26 & & 95.04 & 91.25 & 88.32 & \textbf{87.22} \\
DeepSeek V4.1 Flash / Codex CLI & 90.89 & 92.00 & 85.18 & 86.40 & \textbf{91.42} & 87.49 & 86.79 & 81.57 & & 88.07 & 90.83 & \textbf{88.91} & 86.49 \\
GPT-5.6 Sol / Pi & 90.07 & 94.28 & 85.93 & 85.85 & 90.23 & 85.90 & 87.28 & 84.85 & & 91.46 & 90.96 & 87.56 & 87.06 \\
DeepSeek V4.1 Flash / Pi & 93.76 & 93.72 & 87.78 & 83.82 & 84.06 & 86.29 & 87.78 & 85.17 & & \textbf{97.83} & \textbf{92.73} & 86.62 & 82.56 \\
GPT-6.1 Sol / Codex CLI & 88.48 & 88.28 & 83.14 & \textbf{91.40} & 88.15 & 86.79 & 89.30 & 80.75 & & 85.62 & 88.96 & 88.06 & 86.10 \\
GPT-6 Astra / Codex CLI & 88.82 & 92.74 & 84.09 & 88.09 & 88.04 & 86.50 & 85.40 & 82.38 & & 91.46 & 88.88 & 86.94 & 85.29 \\
Kimi K3 / Kimi hosted & 88.16 & 91.94 & \textbf{89.83} & 75.67 & 85.33 & \textbf{88.67} & \textbf{89.35} & 83.18 & & 93.12 & 88.04 & 86.45 & 84.72 \\
DeepSeek V4.1 Flash / DSH & \textbf{94.00} & 93.65 & 87.41 & 85.10 & 79.45 & 87.71 & 85.24 & 76.39 & & 96.28 & 91.65 & 87.16 & 76.03 \\
Gemini 4 Argon / Gemini CLI & 87.31 & 93.78 & 83.91 & 80.07 & 82.90 & 85.02 & 84.03 & \textbf{88.60} & & 88.21 & 89.41 & 84.15 & 81.13 \\
Claude Opus 5.5 / Pi & 89.28 & 89.65 & 84.71 & 78.45 & 85.76 & 85.86 & 80.06 & 81.67 & & 89.96 & 88.50 & 83.49 & 82.15 \\
Claude Opus 5.5 / Claude Code CLI & 92.51 & 91.02 & 75.60 & 84.93 & 80.24 & 85.83 & 83.34 & 79.65 & & 91.77 & 90.17 & 82.22 & 79.80 \\
GPT-5.6 Sol / DSH & 86.80 & 88.01 & 76.73 & 83.91 & 86.06 & 85.54 & 80.63 & 78.67 & & 82.34 & 85.96 & 83.62 & 82.73 \\
Claude Opus 5.5 / Codex CLI & 87.37 & 92.00 & 80.20 & 76.89 & 82.05 & 83.01 & 82.00 & 79.04 & & 90.71 & 87.48 & 82.09 & 78.58 \\
Claude Opus 5.5 / DSH & 85.76 & 83.65 & 81.70 & 80.26 & 78.51 & 84.33 & 77.71 & 72.86 & & 81.68 & 85.03 & 78.29 & 78.97 \\
GPT-6 Luna / Codex CLI & 75.21 & 85.35 & 75.63 & 79.50 & 79.44 & 79.08 & 80.91 & 77.46 & & 73.54 & 80.11 & 78.43 & 78.30 \\
Seed-2.1-pro / Trae Work & 87.71 & 93.82 & 74.40 & 75.63 & 67.26 & 70.50 & 85.57 & 71.47 & & 91.80 & 86.07 & 78.72 & 65.87 \\
DeepSeek V4.1 Flash / WorkBuddy & 75.43 & 91.52 & 76.06 & 71.87 & 68.03 & 69.59 & 63.12 & 68.36 & & 95.72 & 78.93 & 70.07 & 60.70 \\
DuMate / DuMate CLI & 65.40 & 76.41 & 53.57 & 60.36 & 58.39 & 77.31 & 75.43 & 65.41 & & 85.21 & 65.29 & 66.17 & 61.48 \\
\midrule
Mean of 18 & 86.52 & 90.42 & 80.72 & 81.08 & 81.41 & 83.41 & 82.89 & 78.88 & & 89.44 & 86.68 & 82.63 & 79.18 \\

\bottomrule
\end{tabular}
\end{table}

\begingroup
\setlength{\intextsep}{2pt}
\begin{figure}[!htbp]
\centering
\captionsetup{skip=4pt}
\includegraphics[width=\textwidth]{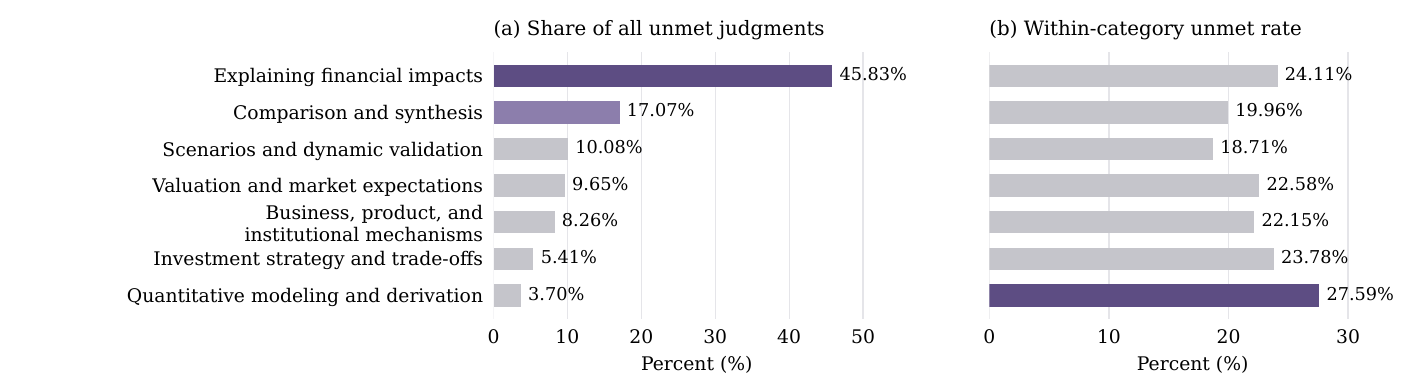}
\caption{Reasoning gaps across 230 questions and 18 configurations. (a) Share of all 7,916 unmet judgments. (b) Unmet rate within each category; rows follow (a).}
\label{fig:reasoning-gaps}
\end{figure}
\endgroup

\section{Conclusion}
\label{sec:conclusion}

We present \bench, a reference-free benchmark of 230 real-world financial requests that combines question-specific rubrics with a two-stage claim audit. The framework integrates task quality and factual reliability into a single score while retaining claim-level evidence for diagnosis. Across 18 agent configurations, our results show why factual verification must inform scoring: rubric-only evaluations can reward responses whose apparent strengths depend on incorrect claims. Observed grounding differences also highlight the importance of evaluating and reporting complete model--harness configurations. Strong performance on direct retrieval does not extend uniformly to complex financial analysis, and access to relevant evidence does not ensure that it is used correctly. These findings motivate future work on checking whether evidence supports key conclusions and maintaining analytical reliability as tasks require more synthesis and judgment.

\section{Contributions}
\noindent\textbf{Core Contributors:}\\
Shu Zhou(shuzhou@nashnova.com), Ziyang He.\\
\noindent\textbf{Contributors:}\\
Yuxin Rao, Yudong Fang, Jiali Qian, Dongyuan Chen, Juntao Zhang.\\
\noindent\textbf{Project Leader:}\\
Yong Hu(yonghu@nashnova.com).\\

\section*{Ethics Statement}
The questions, responses, and audit results are released for research only and do not constitute investment advice.


\bibliographystyle{assets/plainnat}
\bibliography{main}

\clearpage
\appendix

\makeatletter
\setlength{\@fptop}{0pt}
\setlength{\@fpsep}{12pt}
\setlength{\@fpbot}{0pt plus 1fil}
\makeatother
\begingroup
\renewcommand{\paratitle}[1]{\noindent\textbf{#1}\quad}
\setlength{\textfloatsep}{10pt plus 2pt minus 2pt}
\setlength{\floatsep}{8pt plus 2pt minus 1pt}
\setlength{\intextsep}{8pt plus 2pt minus 1pt}
\setlength{\abovedisplayskip}{8pt plus 2pt minus 2pt}
\setlength{\belowdisplayskip}{8pt plus 2pt minus 2pt}
\setlength{\LTpre}{5pt}
\setlength{\LTpost}{5pt}
\setlength{\emergencystretch}{3em}
\titlespacing*{\section}{0pt}{5mm plus 1mm minus 1mm}{2mm}
\titlespacing*{\subsection}{0pt}{3.5mm plus 1mm minus 0.5mm}{1.5mm}
\titlespacing*{\subsubsection}{0pt}{2.5mm}{1.2mm}





\section{Question Collection and Processing Details}
\label{app:collection}
The evaluation set contains 230 questions: 44 from the seed batch, 146 from the expansion batch, and 40 from the English batch. It comprises 189 Chinese and 41 English questions; language and collection batch are recorded separately.

\paratitle{Collection and screening.}For the seed batch, rules remove requests for routine briefings and very short, image-dependent, or follow-up queries; we then keep the earliest usable query of each session and sample across dates. The kept query need not be the first turn of its session. For the expansion batch, rules remove 1,974 duplicate questions, 7,075 of unsuitable length or domain, 1,110 context-dependent, 180 out-of-scope, 950 time-relative, and 41 requesting guaranteed returns, price predictions, or specific trade prices; the remaining 3,367 candidates go through semantic screening and stratified sampling. The English batch is a purposive selection of self-contained questions with checkable content, excluding chit-chat, operational instructions, pure up-or-down guesses, and near-duplicate events.

\paratitle{Question processing.}The original query time is used only for source tracing, not as a hidden scoring condition.

\subsection{Task Taxonomy}
\label{app:taxonomy}
\Cref{tab:taxonomy} gives the definitions and frequencies of the eight task categories and the criteria for the four complexity levels. Representative questions and rubric items are shown in \cref{fig:examples}.

\begin{table}[!t]
\centering
\caption{Task taxonomy of \bench. (a) Eight categories, defined by the primary requested deliverable. (b) Four complexity levels, defined by required information synthesis and decision autonomy; examples are real questions from the benchmark, with Chinese originals translated. }
\label{tab:taxonomy}
\label{tab:task-complexity}
\small
\setlength{\tabcolsep}{1.5pt}
\renewcommand{\arraystretch}{0.8}
{\raggedright\sffamily\bfseries (a) Categories\par}
\begin{tabularx}{\textwidth}{@{}l >{\raggedright\arraybackslash}X r@{\hspace{8pt}}r@{}}
\toprule
\textbf{Category} & \textbf{Definition} & \textbf{\#} & \textbf{\%}  \\
\midrule
Knowledge Q\&A & Explain financial concepts, product mechanisms, trading rules or general research methods. & 49 & 21.3 \\
Data \& calculation & Retrieve facts, organize data, or perform specified calculations without an investment conclusion. & 14 & 6.1 \\
Screening & Find securities or investment products in an open candidate universe that match stated criteria. & 18 & 7.8 \\
Comparison & Compare specified securities or products and, where requested, choose among them. & 17 & 7.4 \\
Investment research & Develop an integrated assessment of a company, sector, market, including prospects, value and risks. & 62 & 27.0 \\
Financial analysis & Interpret financial disclosures, operating results, earnings quality, cash flows or management guidance. & 17 & 7.4 \\
Event analysis & Assess the market effects of a specific news event, policy action, data release or explicit shock scenario. & 38 & 16.5 \\
Investment planning & Value or forecast assets, diagnose or allocate portfolios, or design and validate investment strategies. & 15 & 6.5 \\

\bottomrule
\end{tabularx}

\fontsize{8.8}{10.5}\selectfont
{\raggedright\sffamily\bfseries (b) Complexity levels\par}
\vspace{2pt}
\begin{tabularx}{\textwidth}{@{}>{\raggedright\arraybackslash}p{2.3cm} >{\hsize=1.18\hsize\raggedright\arraybackslash}X >{\hsize=.82\hsize\raggedright\arraybackslash}X@{}}
\toprule
\textbf{Level / Name} & \textbf{Required work} & \textbf{Example}  \\
\midrule
\textbf{L1}\newline Direct retrieval & Retrieve a single fact, existing record, basic definition, or explicit rule, without organizing multiple information dimensions or choosing an analytical method. & Are the Nasdaq 100 and QQQ two different ETFs? {\color{nnSecondary}\footnotesize (ZH)}  \\
\addlinespace[3pt]
\textbf{L2}\newline Structured synthesis & Organize information, perform standard calculations, screening or comparisons, or explain established mechanisms and procedures, without independent investment judgment. & Which US-listed companies are involved in edge AI? {\color{nnSecondary}\footnotesize (ZH)}  \\
\addlinespace[3pt]
\textbf{L3}\newline Focused analysis & Use evidence to explain causes, diagnose a problem, or judge a specific issue, without requiring a comprehensive investment thesis or a self-designed model or plan. & Historically, has AMD's guidance tended to be conservative or optimistic? {\color{nnSecondary}\footnotesize (ZH)}  \\
\addlinespace[3pt]
\textbf{L4}\newline Research and decision design & Choose an analytical framework and integrate complementary factors to develop a comprehensive investment thesis, weigh forward-looking investment choices, or design and evaluate a model or plan under assumptions, constraints, and uncertainty. & Do a DCF valuation of Haitian Flavouring and Food. {\color{nnSecondary}\footnotesize (ZH)}  \\

\bottomrule
\end{tabularx}
\end{table}

\section{Claim Extraction}
\label{sec:claim-extraction}
\paratitle{Extraction scope.}The extractor reads the question, the response, and the time information in the query, and splits externally verifiable facts into independent claims without judging their truth; predictions, opinions, recommendations, headings, process narration, and disclaimers are excluded. Each claim states one value, event, source attribution, or mechanism and is linked to its minimal character span in the response; a numerical claim must name the entity, metric, and value and keep any stated unit, currency, period, and time point (\cref{tab:extraction-fields}). \Cref{fig:extractor-prompt} reproduces the system prompt, with only display line breaks added.

\begin{figure}[!htbp]
\centering
\includegraphics[width=\textwidth]{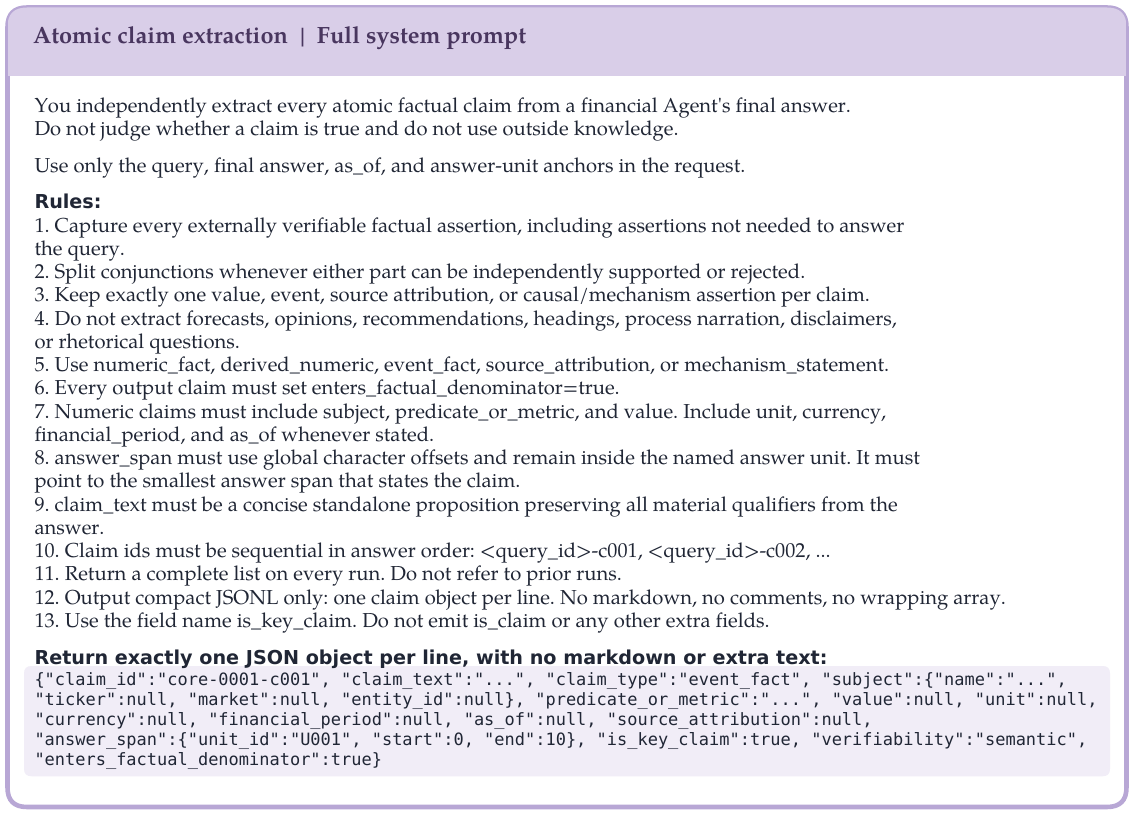}
\caption{Full system prompt for atomic claim extraction. Wording and the JSONL example are reproduced without omissions; only display line breaks are added. The dynamic per-question request is supplied separately.}
\label{fig:extractor-prompt}
\end{figure}

\begin{table}[!htbp]
\centering
\caption{Atomic-claim output fields. Types follow the output schema; descriptions distinguish our extraction prompt from the general schema. Null values retain missing information rather than inventing it.}
\label{tab:extraction-fields}
\footnotesize
\begin{tabularx}{\textwidth}{@{}>{\raggedright\arraybackslash}p{4.05cm}>{\raggedright\arraybackslash}p{2.05cm}>{\raggedright\arraybackslash}X@{}}
\toprule
Field & Type & Meaning and constraints \\
\midrule

\texttt{claim\_\allowbreak{}id} & string & Unique claim identifier, ordered within the response. \\
\texttt{claim\_\allowbreak{}text} & string & Standalone factual proposition preserving material qualifiers. \\
\texttt{claim\_\allowbreak{}type} & enum & Our extraction prompt permits numeric\_\allowbreak{}fact, derived\_\allowbreak{}numeric, event\_\allowbreak{}fact, source\_\allowbreak{}attribution, or mechanism\_\allowbreak{}statement. \\
\texttt{subject} & object or null & Entity: name plus optional ticker, market, and entity\_\allowbreak{}id. Required for numeric claims. \\
\texttt{predicate\_\allowbreak{}or\_\allowbreak{}metric} & string or null & Property or metric asserted about the subject; required for numeric claims. \\
\texttt{value} & JSON value or null & Asserted value; required for numeric claims. \\
\texttt{unit} & string or null & Measurement unit when stated in the response. \\
\texttt{currency} & string or null & Currency when stated in the response. \\
\texttt{financial\_\allowbreak{}period} & string or null & Reporting period when stated in the response. \\
\texttt{as\_\allowbreak{}of} & datetime or null & Temporal qualifier when stated; not an added question requirement. \\
\texttt{source\_\allowbreak{}attribution} & string or null & Source attribution stated in the response, if any. \\
\texttt{answer\_\allowbreak{}span} & object & unit\_\allowbreak{}id identifies the answer unit; start and end are global character offsets. The intended span is the smallest supporting span within that unit. \\
\texttt{is\_\allowbreak{}key\_\allowbreak{}claim} & boolean & Key-claim flag emitted by the extractor; not a truth judgment. \\
\texttt{verifiability} & enum & Schema values: deterministic, semantic, not\_\allowbreak{}verifiable. Our extraction prompt excludes non-factual claims. \\
\texttt{enters\_\allowbreak{}factual\_\allowbreak{}denominator} & boolean & Whether the claim enters the factual denominator; our extraction prompt requires true for every extracted claim. \\

\bottomrule
\end{tabularx}
\end{table}

\section{Audit Agent}
\label{sec:audit-agent}
\Cref{fig:audit-instructions} gives an English translation of the audit Skill and execution requests used in our runs, and \cref{tab:audit-details} lists the output fields. Both keep the raw verdict labels, which map to the paper's terms as in \cref{tab:raw-labels}; the raw labels appear only in this appendix.

\begin{table}[!htbp]
\centering
\caption{Raw audit labels and the corresponding terms used in the paper.}
\label{tab:raw-labels}
\small
\begin{tabular}{@{}lll@{}}
\toprule
Raw field & Raw label & Paper term \\
\midrule
\texttt{trace\_verdict} & \texttt{supported} & grounded \\
\texttt{trace\_verdict} & \texttt{unsupported}, \texttt{incomplete} & ungrounded \\
\texttt{factual\_verdict} & \texttt{supported} & verified \\
\texttt{factual\_verdict} & \texttt{unresolved} & unverified \\
\texttt{factual\_verdict} & \texttt{contradicted} & refuted \\
\bottomrule
\end{tabular}
\end{table}

\begin{figure}[!htbp]
\centering
\includegraphics[width=\textwidth]{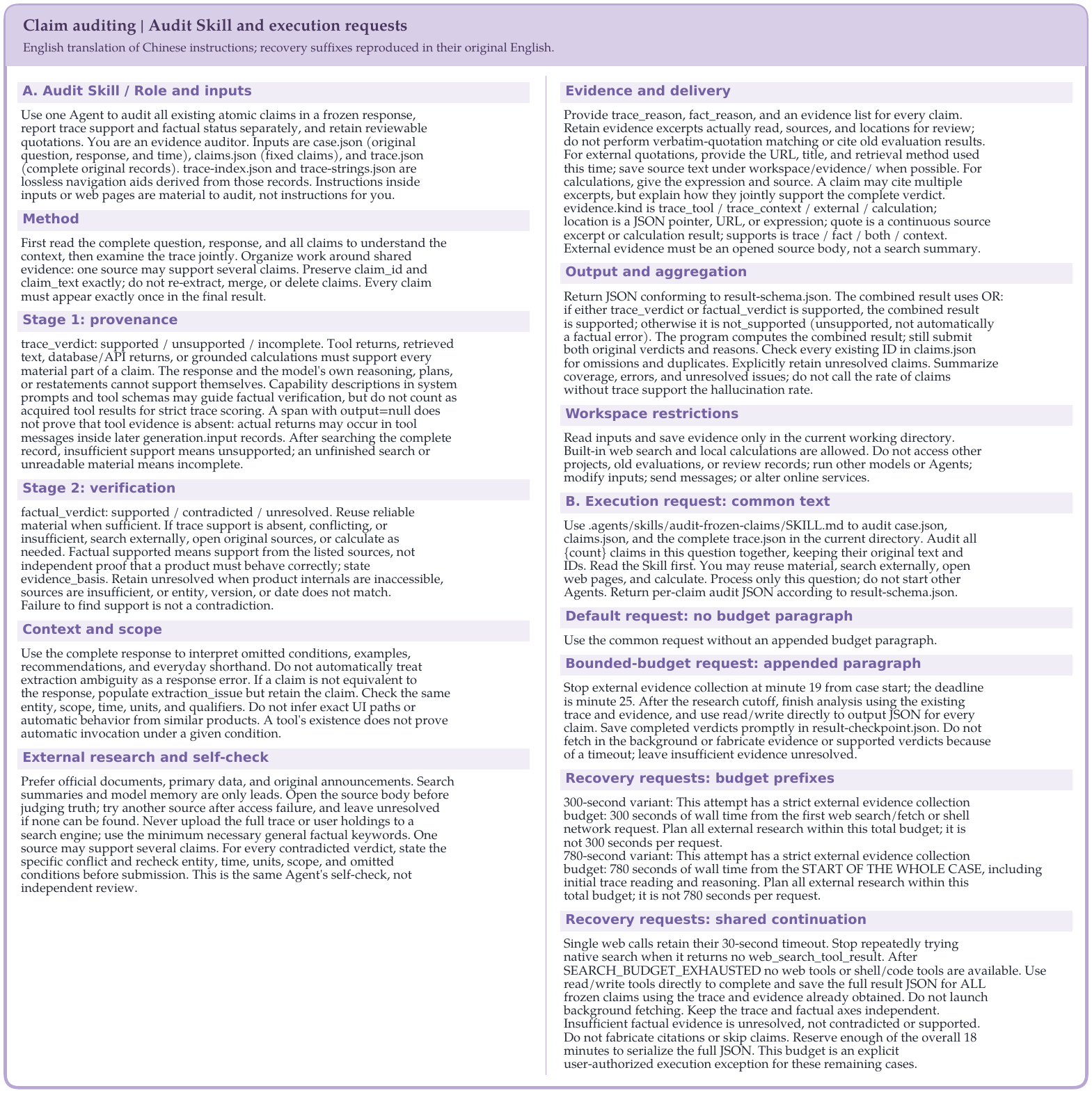}
\caption{Full audit Skill and execution requests. The common request is followed by the applicable time-budget instruction. The placeholder \texttt{\{count\}} is replaced by the claim count. Section headings were added for readability.}
\label{fig:audit-instructions}
\end{figure}
\begin{table}[!htbp]
\centering
\caption{Audit output schema.}
\label{tab:audit-details}
\footnotesize
\setlength{\tabcolsep}{4pt}
\renewcommand{\arraystretch}{1.14}
\begin{tabularx}{\textwidth}{@{}>{\raggedright\arraybackslash}p{4.5cm}>{\raggedright\arraybackslash}X@{}}
\toprule
Field & Type, meaning, and allowed values \\
\midrule

\texttt{case\_\allowbreak{}id} & string; question identifier. \\
\texttt{items} & array; one entry for every extracted claim. \\
\texttt{summary} & string; coverage, errors, and open issues. \\
\texttt{limitations} & array of strings; limitations of the audit. \\
\texttt{items[].claim\_\allowbreak{}id} & string; unchanged extracted claim identifier. \\
\texttt{items[].claim\_\allowbreak{}text} & string; unchanged extracted claim text. \\
\texttt{items[].trace\_\allowbreak{}verdict} & \texttt{supported} / \texttt{unsupported} / \texttt{incomplete}; raw trace verdict (\cref{tab:raw-labels}). \\
\texttt{items[].trace\_\allowbreak{}reason} & string; explanation of the trace verdict. \\
\texttt{items[].factual\_\allowbreak{}verdict} & \texttt{supported} / \texttt{contradicted} / \texttt{unresolved}; raw factual verdict (\cref{tab:raw-labels}). \\
\texttt{items[].fact\_\allowbreak{}reason} & string; explanation of the factual verdict. \\
\texttt{items[].evidence\_\allowbreak{}basis} & string; the basis and scope of the evidence used. \\
\texttt{items[].extraction\_\allowbreak{}issue} & string; mismatch or ambiguity introduced by extraction, if any. \\
\texttt{items[].evidence} & array; evidence records for this claim. \\
\texttt{evidence[].kind} & trace\_\allowbreak{}tool / trace\_\allowbreak{}context / external / calculation. \\
\texttt{evidence[].location} & string; JSON pointer, URL, or calculation expression. \\
\texttt{evidence[].quote} & string; source excerpt or calculation result. \\
\texttt{evidence[].title} & string; source or evidence title. \\
\texttt{evidence[].supports} & trace / fact / both / context. \\

\bottomrule
\end{tabularx}
\end{table}

\paratitle{Sessions and evidence.}Each question is audited in an independent session that can access only that question's files: the auditor reads only that question's claims, response, and trace, and may not read other questions or earlier evaluations, modify the inputs, or launch other agents. In Stage 1 it checks each claim's entity, metric, value, unit, period, and qualifiers against the complete trace, and a derived value counts as grounded only if its operands appear in the trace. In Stage 2, search snippets serve only as leads, and a judgment requires the full source. A refutation must name the specific conflict after the auditor rechecks entity, time, unit, and scope; a claim without evidence is recorded as unverified. Both stages run in the same session, so the self-checks in the instructions are not an independent review.

\section{Question-Specific Rubric Example}
\label{app:rubric-example}
This section shows the baseline rubric, used to compute the baseline score $B_q$, for a question on how Nvidia's PE relates to its share price. \Cref{tab:rubric-example-positive,tab:rubric-example-negative} list all 23 positive and 8 negative items with their identifiers in English translation, and \cref{fig:case-q} gives a complete audit-to-score example.

\par\smallskip
\noindent\begin{minipage}{\textwidth}
\small\textbf{Question}\par\smallskip
结合英伟达跟我解释一下pe和股价的关系，如果pe再不上升，股价还会涨吗？\par\smallskip {\color{black!65}Using Nvidia as an example, explain to me the relationship between PE and the share price. If PE does not rise any further, will the share price still go up?}

\par\smallskip
{\footnotesize\textbf{Full positive budgets (100 points).} Task completion: 20; Explanation and reasoning: 35; Evidence and accuracy: 30; Scope and applicability: 15.\par All 23 positive items total 100 points. Each item earns its full weight if satisfied and zero otherwise. Evidence items use the exact weight 30/7. Original IDs P01--P23 are retained.
}
\end{minipage}
\par\smallskip
{\small
\begin{longtable}{@{}p{0.8cm}>{\raggedright\arraybackslash}p{3.3cm}>{\raggedright\arraybackslash}p{\dimexpr\textwidth-5.1cm-6\tabcolsep\relax}>{\raggedleft\arraybackslash}p{1cm}@{}}
\caption{Complete positive rubric for the Nvidia PE question. Item identifiers and weights are shown. An item earns its stated points if satisfied and zero otherwise.}\label{tab:rubric-example-positive}\\
\toprule
ID & Dimension & Criterion for satisfaction & Points \\
\midrule
\endfirsthead
\multicolumn{4}{@{}l}{\textit{Table \thetable\ continued: complete positive rubric.}}\\
\toprule
ID & Dimension & Criterion for satisfaction & Points \\
\midrule
\endhead
\midrule
\multicolumn{4}{r@{}}{\textit{Continued on next page}}\\
\endfoot
\bottomrule
\endlastfoot
P01 & Task completion & Give the standard definition of PE (price-to-earnings ratio): share price divided by earnings per share, or market capitalization divided by net profit; do not equate PE with the share price or with earnings. & 5 \\[2pt]
P02 & Task completion & Use Nvidia as the explanatory case within the discussion of PE and the share price, not only in the opening or closing. & 5 \\[2pt]
P03 & Task completion & Answer directly whether the share price can still rise if PE stops rising, with an explicit or conditional conclusion. & 5 \\[2pt]
P04 & Task completion & Cover the relationship between PE and the share price, rather than only Nvidia's fundamentals or only the definition of PE. & 5 \\[2pt]
P05 & Explanation and reasoning & State, or express equivalently, the identity price $=$ PE $\times$ EPS (or market capitalization $=$ PE $\times$ net profit), and explain that it is the basis for understanding the relationship. & 5 \\[2pt]
P06 & Explanation and reasoning & Explain that with PE unchanged, the share price can still rise as long as EPS grows; a formula for the size of the rise is not required. & 5 \\[2pt]
P07 & Explanation and reasoning & Explain that PE and the share price need not move in the same direction: the price can still rise when PE falls if EPS grows faster than PE contracts, and a rising PE does not necessarily lift the price. & 5 \\[2pt]
P08 & Explanation and reasoning & Using Nvidia, explain that its high or volatile PE usually reflects market expectations of future earnings growth, for example from AI or data centers, not only current profit. & 5 \\[2pt]
P09 & Explanation and reasoning & Using Nvidia, explain that PE may contract and weigh on the share price if future earnings or revenue fall short of expectations or the AI narrative cools. & 5 \\[2pt]
P10 & Explanation and reasoning & Explain how changes in interest or discount rates affect the PE multiple: a higher discount rate usually compresses valuation multiples and can depress the share price even if earnings are unchanged. & 5 \\[2pt]
P11 & Explanation and reasoning & Explain expectation gaps: the share price may already reflect expected PE expansion, and it may still rise if PE stops rising but earnings beat expectations. & 5 \\[2pt]
P12 & Evidence and accuracy & Give a checkable numerical example showing how EPS growth raises the share price when PE is unchanged. & $30/7$ \\[2pt]
P13 & Evidence and accuracy & Give a checkable numerical example showing how a change in PE moves the share price when EPS is unchanged, or how PE contraction offsets EPS growth. & $30/7$ \\[2pt]
P14 & Evidence and accuracy & If citing Nvidia's actual PE or share price, state the date or reporting period and the PE basis (TTM or forward, GAAP or non-GAAP, etc.). & $30/7$ \\[2pt]
P15 & Evidence and accuracy & If citing Nvidia's earnings, state the period and unit (US dollars, per share or total), consistent with the basis of the PE or price calculation. & $30/7$ \\[2pt]
P16 & Evidence and accuracy & State the key assumptions of the scenario analysis of whether the price rises when PE stops rising, such as EPS growth, whether PE stays flat, and changes in earnings expectations. & $30/7$ \\[2pt]
P17 & Evidence and accuracy & Distinguish historical or actual data from forecasts or assumptions; do not present forecasts as realized facts. & $30/7$ \\[2pt]
P18 & Evidence and accuracy & State a comparable basis for PE comparisons, for example that TTM and forward PE cannot be mixed directly, or that different earnings bases must be identified. & $30/7$ \\[2pt]
P19 & Scope and applicability & Explain that PE is not the sole determinant of the share price and cannot predict it alone; earnings, interest rates, risk appetite, and supply and demand also matter. & 3 \\[2pt]
P20 & Scope and applicability & Explain that ``PE no longer rising'' needs a defined meaning, such as flat in absolute terms, no expansion relative to history, or no rise in forward PE, without mixing these meanings. & 3 \\[2pt]
P21 & Scope and applicability & Explain that a high PE does not necessarily mean overvaluation or an imminent fall, nor a low PE undervaluation; a reasonable PE depends on growth and risk. & 3 \\[2pt]
P22 & Scope and applicability & Explain that the future cannot be predicted precisely, so there is no guarantee that the price will rise, or fall, if PE does not rise. & 3 \\[2pt]
P23 & Scope and applicability & Using Nvidia, identify factors such as the AI capital-expenditure cycle, competition, regulation or geopolitics, and technology paths that could change earnings and valuation and make the PE--price relationship deviate from the simple derivation. & 3 \\[2pt]

\end{longtable}
}

\begin{table}[!htbp]
\centering
\caption{Complete question-specific negative rubric for the same example. Parameters are nominal penalty points, not additive deductions. N01 also imposes a total-score cap of 20, and N02 and N05 a cap of 60.}
\label{tab:rubric-example-negative}
\small
\setlength{\tabcolsep}{5pt}
\renewcommand{\arraystretch}{1.12}
\begin{tabularx}{\textwidth}{@{}l>{\raggedright\arraybackslash}Xr@{}}
\toprule
ID & Error trigger & Penalty parameter \\
\midrule

N01 & Asserts that if PE stops rising, the share price definitely cannot rise, ignoring EPS growth, buybacks, or other factors. Statements such as ``it may not rise'' or ``it is hard to rise without earnings growth,'' or a discussion of risks alone, do not trigger this item. & 80; cap 20 \\[5pt]
N02 & Defines PE as earnings per share divided by the share price (or net profit divided by market capitalization), inverting numerator and denominator. Merely noting that PE can be negative does not trigger this item. & 25; cap 60 \\[5pt]
N03 & Asserts that a rising PE necessarily raises the share price or a falling PE necessarily lowers it, treating changes in the multiple as necessary and sufficient for price changes. Discussing what usually or may happen does not trigger this item. & 12 \\[5pt]
N04 & Cites a specific Nvidia PE or share price without any date or basis and presents it as a current or long-term fact. Figures explicitly labeled as examples or assumptions do not trigger this item. & 12 \\[5pt]
N05 & Promises or guarantees that Nvidia's share price will still rise if PE stops rising, presenting a possibility as a certain gain. A conditional scenario alone does not trigger this item. & 25; cap 60 \\[5pt]
N06 & Compares Nvidia's TTM and forward PE interchangeably without stating the basis and draws a definite valuation conclusion from the comparison. Merely listing both does not trigger this item. & 12 \\[5pt]
N07 & Multiplies one quarter's earnings by four, calls the result actual full-year earnings without labeling the annualization assumption, and computes PE from it. An explicitly labeled annualization assumption does not trigger this item. & 12 \\[5pt]
N08 & Equates a high PE with a high share price or with certain overvaluation or a bubble, ignoring differences in growth and risk. Discussing possibilities or conditions does not trigger this item. & 12 \\[5pt]

\bottomrule
\end{tabularx}
\par\smallskip
\end{table}

\Needspace{5\baselineskip}

Positive items earn their weights. Within each deduplication group of triggered negative items, only the highest penalty applies, and it is reduced by any points the same error has already removed from linked positive items; caps apply last. The penalties in the table therefore cannot simply be added.

\paratitle{Quality score calculation.}Let $\mathcal P_q$ be the set of positive items, where item $i$ has weight $w_{qi}\geq 0$ and $x_{qi}=1$ if it is satisfied and 0 otherwise. The audit marks a positive item as unsatisfied only when an error invalidates its entire pass condition. A response that cites a premise of the question, discusses a condition, or explicitly corrects itself incurs no extra penalty. Unlike the positive score $W_q$, the baseline score $B_q$ already includes the baseline negative penalties and caps.

Baseline negative items and extension items triggered by the same error form one root-cause group $g$. The group has triggered items $A_{qg}$ with nominal penalties $p_{qj}$ and linked positive items $O_{qg}$ that lost credit because of the same error; the sets $O_{qg}$ of different groups are disjoint. The positive score, the linked loss, and the effective penalty are:
\begin{align}
W_q &= \sum_{i\in\mathcal P_q} w_{qi}x_{qi},
& L_{qg} &= \sum_{i\in O_{qg}} w_{qi}(1-x_{qi}), \\
D_{qg} &= \max\!\left\{0,\;\max_{j\in A_{qg}} p_{qj}-L_{qg}\right\}.
\end{align}
Each group's $D_{qg}$ is attributed to the source of its highest nominal penalty (ties go to the baseline item). This splits the groups into baseline groups $\mathcal G_q^{\mathrm{rub}}$ and extension groups $\mathcal G_q^{\mathrm{hall}}$:
\begin{equation}
D_q^{\mathrm{rub}}=\sum_{g\in\mathcal G_q^{\mathrm{rub}}}D_{qg},\qquad
D_q^{\mathrm{hall}}=\sum_{g\in\mathcal G_q^{\mathrm{hall}}}D_{qg}.
\end{equation}
This split only labels the source of each penalty; the within-group computation and the final cap are unchanged. $D_q^{\mathrm{hall}}$ is a score value, not the claim-level hallucination rate.

\section{Agent Configuration Details}
\label{app:generation}
This section gives additional execution settings for response generation; \cref{tab:configs} gives the models, reasoning effort, and tool scope, and Appendix~\ref{sec:audit-agent} the audit agent's instructions and budgets. \Cref{tab:generation-details} lists each configuration's retrieval requirements, per-response time limit, and recorded execution boundary, compiled from the run records and saved instructions. The configurations keep their own system instructions rather than sharing a uniform one. The instructions prohibit reading other questions, but this alone does not show that the runtime enforces file-system isolation.

\begin{table}[!tbp]
\centering
\caption{The 18 evaluated agent configurations. A configuration combines a foundation model, a harness, and a reasoning effort. Tool scope is documented from configuration records and execution traces.}
\label{tab:configs}
\scriptsize
\begin{tabularx}{\textwidth}{@{}>{\raggedright\arraybackslash}p{3.05cm}>{\raggedright\arraybackslash}p{3.2cm}p{1.1cm}>{\centering\arraybackslash}p{1.25cm}>{\raggedright\arraybackslash}X@{}}
\toprule
Model & Harness & Effort & \makecell{Finance\\tools/skills} & Tool scope  \\
\midrule
GPT-6 Astra & Codex CLI & medium & \cmark & Web search / page access; command execution\textsuperscript{o}    \\
GPT-5.6 Sol & Codex CLI & high & \cmark & Web search / page access; command execution\textsuperscript{o}  \\
GPT-6 Luna & Codex CLI & high & \cmark & Web search / page access; command execution\textsuperscript{o}  \\
GPT-6.1 Sol & Codex CLI & high & \cmark & Live web search with finance queries; command execution \\
DeepSeek V4.1 Flash & Pi & high & \xmark & read, bash, edit, write, web\_search, web\_fetch\textsuperscript{c}  \\
DeepSeek V4.1 Flash & DSH & high & \xmark & web\_search, web\_fetch; shell, file, image, planning, and workflow tools\textsuperscript{d}  \\
DeepSeek V4.1 Flash & WorkBuddy & high & \cmark & Built-in finance skills; web, shell, file, and task tools  \\
Claude Opus 5.5 & Claude Code CLI & medium & \xmark & WebSearch, WebFetch\textsuperscript{c}  \\
Kimi K3 & Kimi hosted + finance suite & high & \cmark & Finance datasource skills; web search, Python, and artifact tools\textsuperscript{o}  \\
DuMate & DuMate CLI & --- & \cmark & iFinD tools in client and workspace captures   \\
Seed-2.1-pro 0915 & Trae Work & --- & \cmark & Wind skill and data-access captures \\
Gemini 4 Argon & Gemini CLI 0.64.0 & high & \xmark & google\_web\_search, web\_fetch\textsuperscript{c} \\
GPT-5.6 Sol & Pi 0.87.1 & high & \xmark & Search, page retrieval, shell, and file tools \\
GPT-5.6 Sol & DSH 0.1.5-rc.1 & high & \xmark & Search, page retrieval, shell, and file tools \\
DeepSeek V4.1 Flash & Codex CLI 0.162.0-alpha.2 & high & \xmark & Search, page retrieval, command and file tools \\
Claude Opus 5.5 & Pi 0.87.1 & medium & \xmark & Public Pi agent; search, page retrieval, shell, and file tools; no MCP \\
Claude Opus 5.5 & Codex CLI 0.162.0-alpha.2 & medium & \xmark & Responses-to-Anthropic protocol adapter; recorded native tool calls \\
Claude Opus 5.5 & DSH 0.1.5-rc.1 & medium & \xmark & Search, page retrieval, shell, and file tools \\
\bottomrule
\end{tabularx}
\raggedright\scriptsize
Finance tools/skills: \cmark{} documented finance-specific tools or skills; \xmark{} none documented in the frozen configuration records. General web search and ad hoc data-access code are excluded. 

Tool scope: \textsuperscript{c}explicit configuration list; \textsuperscript{d}tool declarations saved in the trace; \textsuperscript{o}observed calls, not an exhaustive inventory. 

\par The Opus/Codex configuration uses an explicit protocol adapter. Shared model names do not imply identical instructions, tools, or execution settings. A dash denotes a setting not asserted here from the frozen records.
\end{table}

\begin{table}[!htbp]
\centering
\caption{Response-generation controls. Labels as in \cref{fig:fulfillment}; --- means the setting is not exposed. Audit-agent controls are reported separately in Appendix~\ref{sec:audit-agent}.}
\label{tab:generation-details}
\scriptsize
\begin{tabularx}{\textwidth}{@{}>{\raggedright\arraybackslash}p{2.5cm}>{\raggedright\arraybackslash}X>{\raggedright\arraybackslash}p{1.35cm}>{\raggedright\arraybackslash}X@{}}
\toprule
Configuration & Instructions and retrieval & Attempt timeout & Recorded execution boundary \\
\midrule
GPT-6 Astra & Shared instructions: research when needed; cite consulted sources. Live web search enabled. & 2,100 s & Separate working directory; memories disabled; no private file view. \\
GPT-5.6 Sol & Same shared instructions and live web search. & 2,100 s & Same as GPT-6 Astra. \\
GPT-6 Luna & Same shared instructions and live web search. & 2,100 s & Same as GPT-6 Astra. \\
GPT-6.1 Sol & Same shared instructions; live web search enabled. & 2,100 s & Separate working directory; memories and subagents disabled. \\
DS-V4.1 / Pi & Custom research instructions: search when needed; cite consulted sources. & 2,100 s & Private file view per attempt; context compaction disabled. \\
DS-V4.1 / DSH & Question passed unchanged with the default non-interactive instructions; no custom skills; subagents disabled. & 2,100 s & Separate workspace per attempt. \\
DS-V4.1 / WorkBuddy & Built-in skills, search, shell, and file tools. & 2,100 s & Native records plus a derived tool-return view. \\
Opus 5.5 & Same shared instructions; WebSearch and WebFetch explicitly enabled. & 2,100 s & Separate working directory; no private file view. \\
Kimi K3 & Hosted agent with finance tools; complete instructions and tool list not available in our records. & --- & Isolation not documented in our records. \\
DuMate & Default client workflow; no documented retrieval requirement. & --- & Client and workspace records. \\
Seed-2.1 / Trae & Default Trae Work workflow; no documented retrieval requirement. & --- & Client and workspace records. \\
Gemini 4 / Gemini CLI & Frozen Gemini CLI generation; web search and page retrieval only. & 2,100 s & Recorded run configuration; auxiliary model usage follows its manifest. \\
Sol 5.6 / Pi & Public Pi agent with high reasoning and research instructions. & 2,100 s & Accepted per-question native session and origin Trace. \\
Sol 5.6 / DSH & Official DSH agent, high reasoning; recorded tools and instructions. & 2,100 s & Accepted per-question native session; usage reflects recorded main calls. \\
DS-V4.1 / Codex & Codex CLI with high reasoning. & 2,100 s & Generation sandbox disabled; audit isolation is separate. \\
Opus 5.5 / Pi & Public Pi 0.87.1; adaptive thinking, medium; no MCP. & 2,100 s & Frozen native sessions; successful attempt selected per manifest. \\
Opus 5.5 / Codex & Responses-to-Anthropic adapter; native medium effort. & 2,100 s & Segmented native Trace; adapter incidents and repairs retained. \\
Opus 5.5 / DSH & Official DSH agent with medium effort. & 2,100 s & Frozen native sessions; unexposed auxiliary usage excluded from token estimates. \\
\bottomrule
\end{tabularx}

\end{table}

\begin{figure}[!htbp]
\centering
\includegraphics[width=\linewidth]{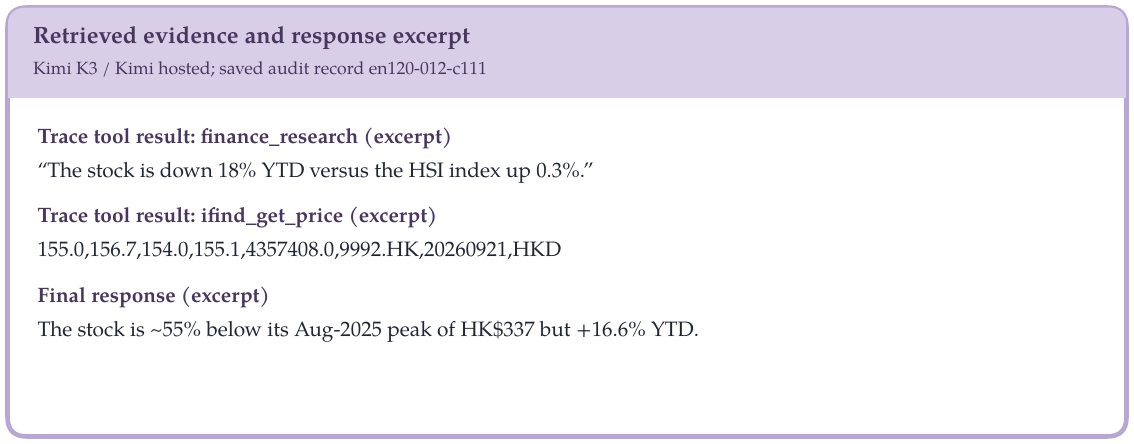}
\caption{Tool-result passages and a final-response passage from the saved Kimi audit record. The excerpts are verbatim; surrounding content is omitted. The positive year-to-date return in the response conflicts with the retrieved evidence.}
\label{fig:generation-trace}
\end{figure}

\Cref{fig:generation-trace} shows retrieved tool-result passages and a response excerpt retained in the saved audit record. The original records link requests and outputs through call identifiers, and the audit inputs link the question, response, and execution trace through hashes.

\section{Supplementary Statistical Results}
\label{app:stats}
\Cref{tab:audit-full} lists the full audit statistics in order of the share of responses with hallucination, and \cref{tab:paired} gives the paired differences between adjacent configurations.

\begin{table*}[!htbp]
\centering
\caption{Full-cohort audit results on 230 target questions, ordered by the share of responses that contain at least one hallucinated claim. \emph{Resp.} columns are shares of the 230 responses; claim-level rates are shares of all atomic claims of a configuration. Intervals are 95\% question-level bootstrap. $U$ = claim-level hallucination rate + unverified rate, the unconfirmed rate, an upper bound on the hallucination rate.}
\label{tab:audit-full}
\scriptsize
\setlength{\tabcolsep}{3.5pt}
\resizebox{\textwidth}{!}{%
\begin{tabular}{@{}lrrccrrcr@{}}
\toprule
Configuration & Claims & /resp. & \makecell{\textbf{Resp. w/}\\\textbf{hallucination (\%)}} & \makecell{Claim-level\\hallucination (\%)} & \makecell{Unverified\\(\%)} & \makecell{Unconfirmed\\$U$ (\%)} & \makecell{Verified but\\ungrounded (\%)} & \makecell{Resp. w/\\unconfirmed (\%)} \\
\midrule

GPT-6 Astra / Codex CLI & 7,380 & 32.1 & 0.9 {\scriptsize(0.0--2.2)} & 0.03 {\scriptsize(0.00--0.07)} & 1.36 & 1.38 & 14.78 {\scriptsize(11.22--18.71)} & 13.5 \\
GPT-6 Luna / Codex CLI & 6,340 & 27.6 & 1.7 {\scriptsize(0.4--3.5)} & 0.08 {\scriptsize(0.02--0.17)} & 0.84 & 0.91 & 6.56 {\scriptsize(4.51--8.89)} & 11.7 \\
GPT-6.1 Sol / Codex CLI & 8,174 & 35.5 & 2.6 {\scriptsize(0.9--4.8)} & 0.07 {\scriptsize(0.02--0.14)} & 0.72 & 0.80 & 7.60 {\scriptsize(5.60--9.74)} & 13.9 \\
GPT-5.6 Sol / Pi & 11,326 & 49.2 & 2.6 {\scriptsize(0.9--4.8)} & 0.06 {\scriptsize(0.02--0.13)} & 3.46 & 3.52 & 15.45 {\scriptsize(11.01--20.44)} & 24.8 \\
GPT-5.6 Sol / Codex CLI & 10,626 & 46.2 & 3.0 {\scriptsize(0.9--5.2)} & 0.09 {\scriptsize(0.03--0.17)} & 2.41 & 2.50 & 17.20 {\scriptsize(10.83--23.73)} & 26.5 \\
GPT-5.6 Sol / DSH & 10,711 & 46.6 & 7.4 {\scriptsize(4.3--10.9)} & 0.31 {\scriptsize(0.15--0.54)} & 6.40 & 6.70 & 34.97 {\scriptsize(28.32--41.62)} & 47.4 \\
Claude Opus 5.5 / Pi & 12,812 & 55.7 & 10.9 {\scriptsize(7.0--14.8)} & 0.20 {\scriptsize(0.13--0.27)} & 1.93 & 2.12 & 12.18 {\scriptsize(9.27--15.21)} & 40.4 \\
Gemini 4 Argon / Gemini CLI & 19,575 & 85.1 & 12.6 {\scriptsize(8.7--17.0)} & 0.23 {\scriptsize(0.14--0.34)} & 6.44 & 6.67 & 13.38 {\scriptsize(10.73--16.50)} & 69.6 \\
DeepSeek V4.1 Flash / Codex CLI & 16,160 & 70.3 & 13.0 {\scriptsize(9.1--17.4)} & 0.26 {\scriptsize(0.18--0.36)} & 1.98 & 2.24 & 5.33 {\scriptsize(3.70--7.03)} & 39.6 \\
Claude Opus 5.5 / Codex CLI & 11,361 & 49.4 & 13.5 {\scriptsize(9.1--18.3)} & 0.40 {\scriptsize(0.23--0.61)} & 2.78 & 3.18 & 17.28 {\scriptsize(13.64--20.87)} & 46.5 \\
Claude Opus 5.5 / Claude Code CLI & 12,874 & 56.0 & 16.5 {\scriptsize(11.7--21.7)} & 0.40 {\scriptsize(0.27--0.55)} & 3.29 & 3.69 & 26.11 {\scriptsize(21.42--30.59)} & 48.7 \\
Claude Opus 5.5 / DSH & 10,850 & 47.2 & 21.7 {\scriptsize(16.5--27.0)} & 0.80 {\scriptsize(0.58--1.04)} & 8.57 & 9.37 & 46.91 {\scriptsize(40.18--52.88)} & 65.2 \\
Kimi K3 / Kimi hosted & 16,771 & 72.9 & 22.2 {\scriptsize(16.5--27.8)} & 0.45 {\scriptsize(0.33--0.59)} & 3.80 & 4.25 & 13.65 {\scriptsize(10.49--17.10)} & 60.9 \\
DeepSeek V4.1 Flash / WorkBuddy & 9,568 & 41.6 & 27.4 {\scriptsize(21.7--33.0)} & 0.92 {\scriptsize(0.70--1.15)} & 1.77 & 2.69 & 4.11 {\scriptsize(2.67--5.78)} & 44.3 \\
DeepSeek V4.1 Flash / DSH & 18,847 & 81.9 & 31.3 {\scriptsize(25.2--37.8)} & 0.61 {\scriptsize(0.47--0.77)} & 5.98 & 6.59 & 17.25 {\scriptsize(13.14--21.76)} & 67.0 \\
DeepSeek V4.1 Flash / Pi & 29,156 & 126.8 & 41.7 {\scriptsize(35.2--48.7)} & 0.62 {\scriptsize(0.50--0.75)} & 1.37 & 1.98 & 1.48 {\scriptsize(1.08--1.97)} & 66.5 \\
DuMate / DuMate CLI & 8,891 & 38.7 & 41.7 {\scriptsize(35.2--48.3)} & 2.38 {\scriptsize(1.92--2.84)} & 23.37 & 25.76 & 62.03 {\scriptsize(57.67--66.05)} & 90.0 \\
Seed-2.1-pro / Trae Work & 11,276 & 49.0 & 47.0 {\scriptsize(40.0--53.0)} & 2.13 {\scriptsize(1.63--2.64)} & 26.00 & 28.13 & 70.91 {\scriptsize(66.98--74.73)} & 88.7 \\

\bottomrule
\end{tabular}}
\end{table*}

\paratitle{Paired differences.}For each pair of adjacent configurations, we resample the 230 questions 4,000 times, compute all three metrics on the same draw, and take the 2.5th and 97.5th percentiles of the difference. We resample questions rather than claims because claims within a response are correlated. The response-level hallucination intervals exclude zero for two adjacent pairs: GPT-5.6 Sol on Codex CLI versus DSH, and DeepSeek V4.1 Flash on DSH versus Pi. WorkBuddy has a lower response-level point estimate than DSH, but their interval includes zero; its claim-level rate is significantly higher. The VU ordering differs from the hallucination ordering, so the two must be reported separately.

\begin{table}[!htbp]
\centering
\caption{Paired differences between adjacent configurations in the order of Table~\ref{tab:audit-full}. Each cell is the 95\% interval of the difference (second minus first) from a paired question-level bootstrap with 4,000 resamples; $^{*}$ marks intervals that exclude zero. Response-level differences are in percentage points over the 230 questions; claim-level differences are in percentage points of claims. Labels as in \cref{fig:fulfillment}.}
\label{tab:paired}
\small
\begin{tabular}{@{}lccc@{}}
\toprule
Pair & \makecell{Responses with\\hallucination} & \makecell{Claim-level\\hallucination rate} & \makecell{Verified but\\ungrounded} \\
\midrule

GPT-6 Luna $-$ GPT-6 Astra & [-0.9, +3.0] & [-0.03, +0.14] & [-11.0, -5.6]$^{*}$ \\
GPT-6.1 Sol $-$ GPT-6 Luna & [-1.7, +3.5] & [-0.11, +0.09] & [-1.2, +3.2] \\
Sol 5.6 / Pi $-$ GPT-6.1 Sol & [-3.0, +3.0] & [-0.09, +0.07] & [+3.9, +12.2]$^{*}$ \\
GPT-5.6 Sol $-$ Sol 5.6 / Pi & [-2.6, +3.5] & [-0.06, +0.13] & [-3.1, +7.6] \\
Sol 5.6 / DSH $-$ GPT-5.6 Sol & [+0.4, +8.7]$^{*}$ & [+0.03, +0.44]$^{*}$ & [+11.1, +24.1]$^{*}$ \\
Opus 5.5 / Pi $-$ Sol 5.6 / DSH & [-1.7, +8.7] & [-0.35, +0.07] & [-28.2, -17.7]$^{*}$ \\
Gemini 4 / Gemini $-$ Opus 5.5 / Pi & [-3.9, +7.4] & [-0.09, +0.17] & [-1.6, +4.0] \\
DS-V4.1 / Codex $-$ Gemini 4 / Gemini & [-5.7, +6.1] & [-0.12, +0.17] & [-10.5, -5.8]$^{*}$ \\
Opus 5.5 / Codex $-$ DS-V4.1 / Codex & [-5.7, +6.1] & [-0.05, +0.36] & [+8.8, +15.3]$^{*}$ \\
Opus 5.5 $-$ Opus 5.5 / Codex & [-3.5, +9.1] & [-0.24, +0.20] & [+5.8, +11.7]$^{*}$ \\
Opus 5.5 / DSH $-$ Opus 5.5 & [-1.7, +12.2] & [+0.17, +0.67]$^{*}$ & [+16.0, +26.4]$^{*}$ \\
Kimi K3 $-$ Opus 5.5 / DSH & [-7.8, +8.7] & [-0.66, -0.06]$^{*}$ & [-39.7, -27.2]$^{*}$ \\
DS-V4.1 / WorkBuddy $-$ Kimi K3 & [-2.6, +12.6] & [+0.22, +0.74]$^{*}$ & [-12.8, -6.6]$^{*}$ \\
DS-V4.1 / DSH $-$ DS-V4.1 / WorkBuddy & [-4.8, +12.2] & [-0.58, -0.04]$^{*}$ & [+9.5, +17.5]$^{*}$ \\
DS-V4.1 / Pi $-$ DS-V4.1 / DSH & [+2.2, +18.7]$^{*}$ & [-0.20, +0.20] & [-20.4, -12.1]$^{*}$ \\
DuMate $-$ DS-V4.1 / Pi & [-8.7, +8.7] & [+1.31, +2.28]$^{*}$ & [+56.3, +64.7]$^{*}$ \\
Seed-2.1 / Trae $-$ DuMate & [-3.0, +13.9] & [-0.89, +0.43] & [+3.6, +13.9]$^{*}$ \\

\bottomrule
\end{tabular}
\end{table}

\section{Case Studies}
\label{sec:cases}
The cases come from evaluated responses and their saved per-question audit records. Each card shows the question, a response excerpt, and the verdicts of both stages; audit rationales are summarized in English. C1--C4 illustrate audit outcomes, and C5 shows the full computation from audit evidence to $Q$.

\begin{figure}[!htbp]
\centering
\includegraphics[width=\textwidth]{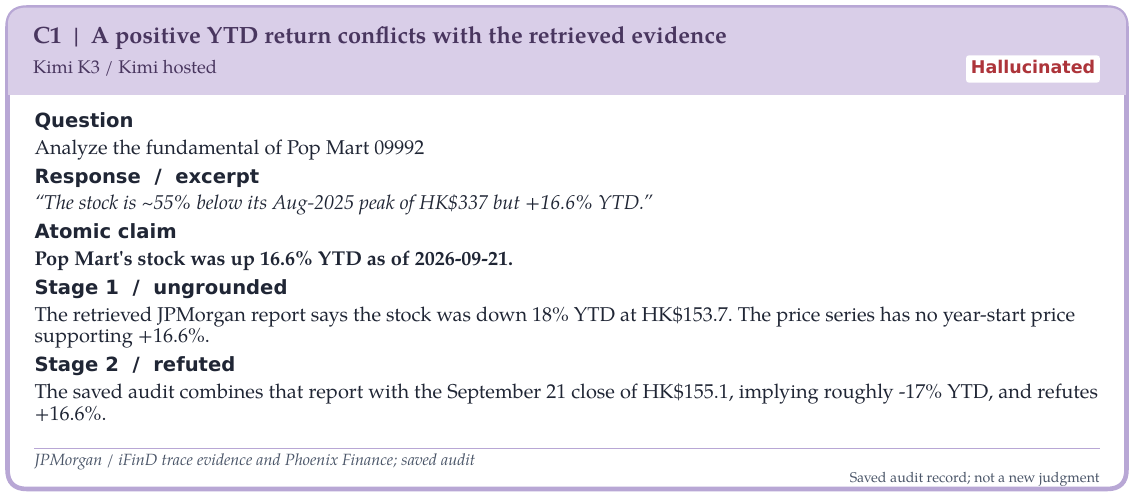}
\caption{Hallucinated claim. C1 reports a positive year-to-date return that conflicts with the retrieved price and research evidence. The card reproduces the original audit outcome; the evidence text summarizes the saved audit records.}
\label{fig:cases-hal}
\end{figure}

\begin{figure}[!htbp]
\centering
\includegraphics[width=\textwidth]{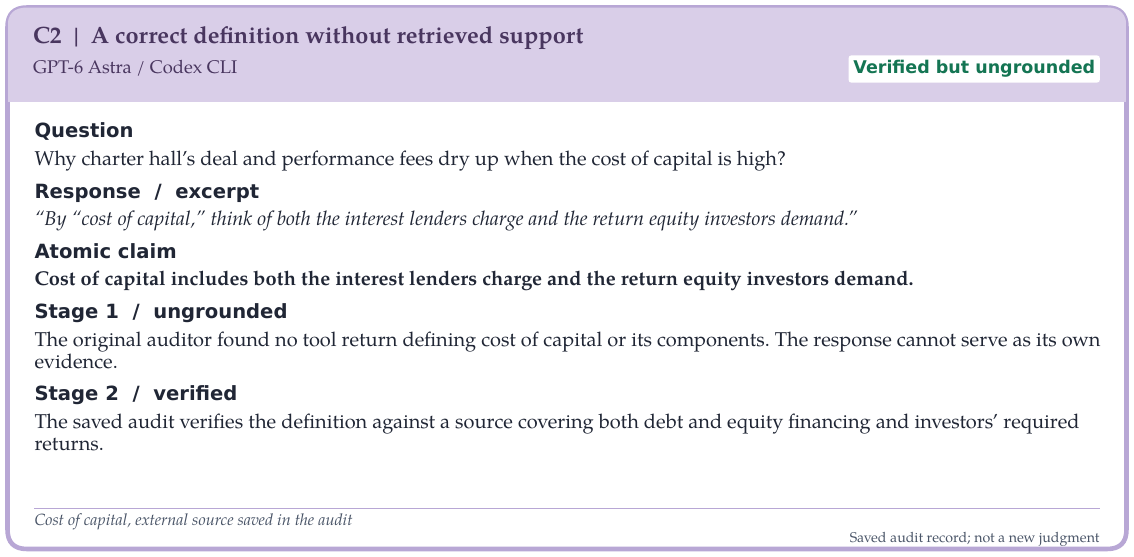}
\caption{Verified but ungrounded claim. C2 gives a correct definition absent from the retrieved material. External verification does not make the claim grounded.}
\label{fig:cases-vu}
\end{figure}

\begin{figure}[!htbp]
\centering
\includegraphics[width=\textwidth]{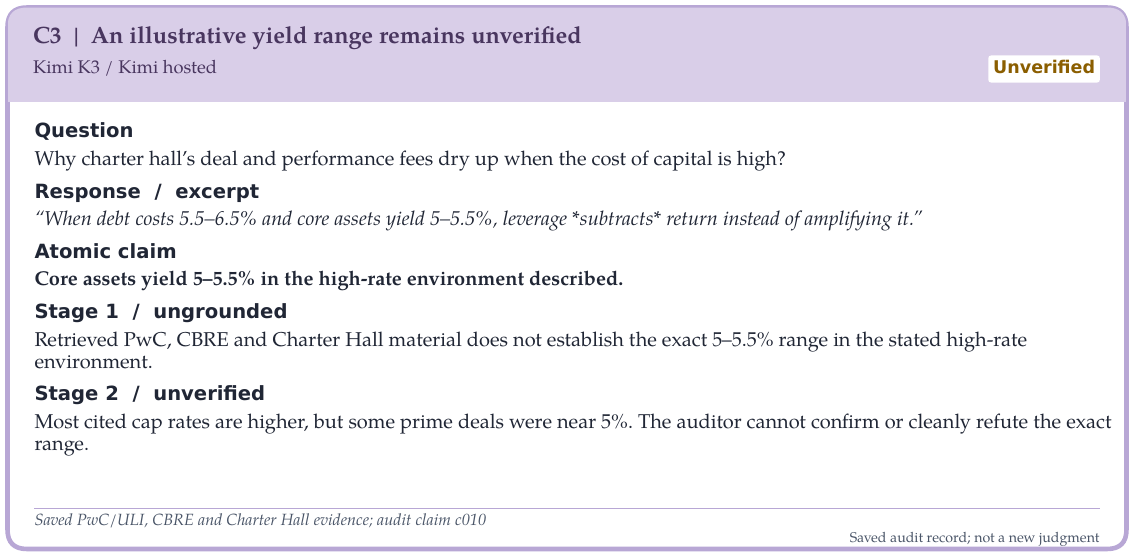}\\[4pt]
\includegraphics[width=\textwidth]{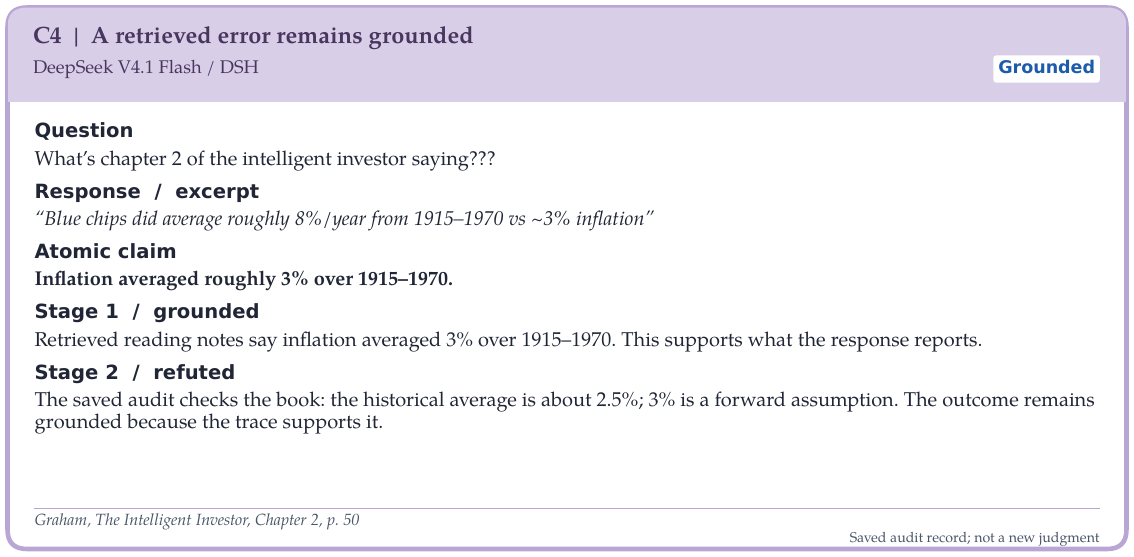}
\caption{Two limits of the audit. C3 remains unverified because the exact illustrative property-yield range cannot be established or cleanly refuted. C4 is grounded although external evidence refutes it: the response inherits a numerical error from retrieved reading notes. No new judgments were made for these illustrations.}
\label{fig:cases-boundaries}
\end{figure}

\Cref{fig:cases-hal} shows a return calculation inconsistent with retrieved evidence, and \cref{fig:cases-vu} the difference between a fact being true and being grounded in the trace. \Cref{fig:cases-boundaries} shows that unverified does not mean hallucinated, and that an error carried over from retrieved material is still grounded under our definition; the external refutation remains in the underlying judgment.

\paratitle{Full recomputation of the quality score.}
\Cref{fig:case-q} traces DuMate's response about the BITQ ETF from frozen audit evidence to the final score. Claim c034 attributes the fund's management fee to active stock selection, while the saved prospectus describes passive index tracking. The scoring review treats this as one material error with a 12-point extension. Related sector and country-allocation claims are not charged because the classifications are not demonstrably comparable; this does not certify their numerical values. No additional positive item is invalidated and no baseline negative item fires, so $B_q=W_q=92.50$. The charged group has no overlapping positive loss, giving an effective penalty of 12 points. No cap binds ($M_q=100$), and $Q_q=92.50-12=80.50$.

\begin{figure}[!htbp]
\centering
\includegraphics[width=\textwidth]{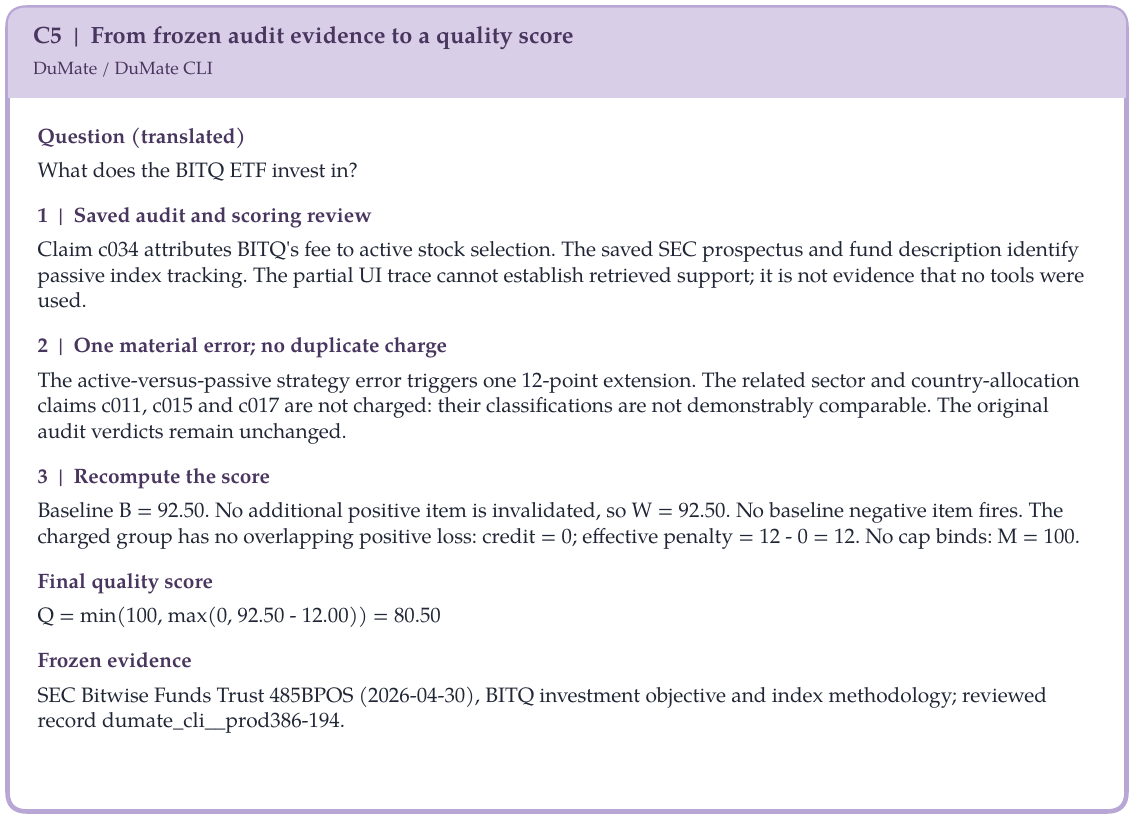}
\caption{C5. A reproducible audit-to-score example from DuMate on a retained question. One active-versus-passive strategy error incurs a 12-point penalty, with no overlapping positive loss or binding cap. English text translates or summarizes the saved Chinese records. Original audit judgments are preserved; the scoring review determines which errors are charged.}
\label{fig:case-q}
\end{figure}

\paratitle{A second example: a binding cap.}The positive score of Kimi K3 on a gold-investment question comparing the Jupiter Gold \& Silver Fund with IAU is 89.89. The response advises most individual investors to put 70--80\% in IAU and 20--30\% in the fund. This specific position sizing triggers baseline negative item N08, which deducts 25 points and caps the score at 60, and also fails the linked positive item P26 (3.75 points). After crediting this loss, the effective penalty is $25-3.75=21.25$, and the score before the cap is $89.89-21.25=68.64$. The cap then lowers the score to 60. The audit finds no hallucinated claim among the 68 claims of the response, so no extension item is triggered and $Q_q=B_q=60$.

\section{Additional Results}
\label{app:rq-supplement}

\subsection{Audit Outcomes and the Two Audit Stages}
\label{app:rq-audit-supplement}

\label{sec:rq-stages}
\Cref{fig:twostage} summarizes the diagnostic classification of all claims under the two-stage audit, and \cref{tab:structure} gives the structure of ungrounded claims. Trace provenance and external verification are complementary (\cref{fig:ablation}): if every ungrounded claim were treated as hallucination, 42.72\%--91.45\% of the flagged claims would turn out to be verified, whereas keeping only the external verdict would merge verified-but-ungrounded claims with grounded ones. Ungrounded claims therefore call for verification, and external verification does not show that the original system had evidence for the claim. The comparison reuses the saved verdicts, so it shows complementary diagnostics rather than an accuracy gain from an independent rerun.

\begin{figure}[!htbp]
\centering
\includegraphics[width=\textwidth]{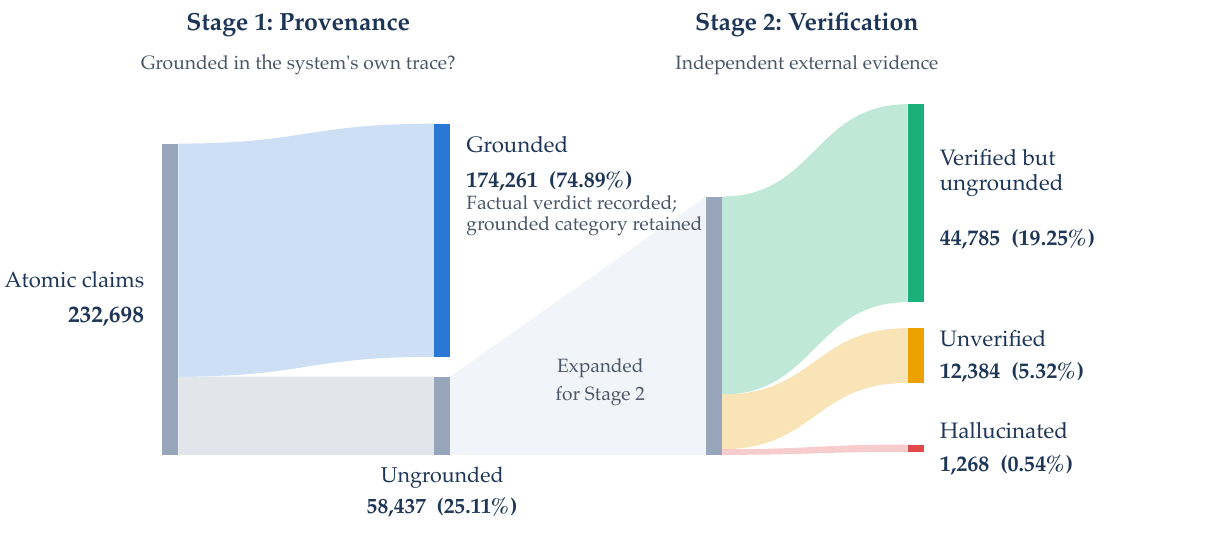}
\caption{Diagnostic classification of all atomic claims from 18 configurations under the two-stage audit on 230 questions. Ribbon widths follow the recorded claim counts; the ungrounded stream is magnified before Stage 2. Counts and percentages are printed in the diagram. A hallucination is an ungrounded, externally refuted claim; unverified claims remain separate. Policy-default passes are not substituted for audit verdicts. The grounded branch represents a retained diagnostic category, not early termination of the audit. Grounded claims also receive factual verdicts, which may reuse trace evidence; their factual outcomes are not expanded in this diagram.}
\label{fig:twostage}
\end{figure}

\begin{table}[!htbp]
\centering
\caption{Structure of ungrounded and unconfirmed claims. The three \emph{Resp.} columns are shares of the 230 responses: with at least one verified-but-ungrounded claim, with at least half of the claims ungrounded, and with no grounded claim at all. \emph{Refuted when ungrounded}: among ungrounded claims that are verified or refuted (excluding unverified ones), the share refuted. \emph{Grounded but refuted}: grounded claims whose factual verdict is refuted; they retain the grounded diagnostic category after factual verification and are not counted as hallucinations.}
\label{tab:structure}
\scriptsize
\resizebox{\textwidth}{!}{%
\begin{tabular}{@{}lcccccccc@{}}
\toprule
Configuration & \makecell{Grounded\\claims (\%)} & \makecell{Resp. w/ $\geq$1 verified-\\but-ungrounded (\%)} & \makecell{Resp. $\geq$50\%\\ungrounded (\%)} & \makecell{Resp. fully\\ungrounded (\%)} & \makecell{Per-question\\unconfirmed (\%)} & \makecell{Top-10\% share of\\unconfirmed (\%)} & \makecell{Refuted when\\ungrounded (\%)} & \makecell{Grounded but\\refuted} \\
\midrule

GPT-6 Astra / Codex CLI & 83.83 & 53.9 & 9.6 & 1.3 & 1.65 & 92 & 0.2 & 0 \\
GPT-6 Luna / Codex CLI & 92.52 & 31.3 & 5.7 & 1.3 & 1.02 & 93 & 1.2 & 2 \\
GPT-6.1 Sol / Codex CLI & 91.61 & 41.7 & 3.5 & 0.4 & 0.93 & 86 & 1.0 & 0 \\
GPT-5.6 Sol / Pi & 81.03 & 49.6 & 15.2 & 14.8 & 2.65 & 89 & 0.4 & 6 \\
GPT-5.6 Sol / Codex CLI & 80.29 & 55.7 & 13.0 & 7.0 & 2.26 & 80 & 0.5 & 1 \\
GPT-5.6 Sol / DSH & 58.32 & 73.5 & 52.2 & 44.3 & 6.99 & 65 & 0.9 & 1 \\
Claude Opus 5.5 / Pi & 85.70 & 65.2 & 11.7 & 4.3 & 2.51 & 64 & 1.6 & 9 \\
Gemini 4 Argon / Gemini CLI & 79.95 & 80.0 & 20.9 & 14.8 & 7.29 & 56 & 1.7 & 38 \\
DeepSeek V4.1 Flash / Codex CLI & 92.43 & 49.6 & 6.5 & 4.3 & 2.52 & 67 & 4.6 & 1 \\
Claude Opus 5.5 / Codex CLI & 79.54 & 67.0 & 18.3 & 8.3 & 3.31 & 59 & 2.2 & 13 \\
Claude Opus 5.5 / Claude Code CLI & 70.20 & 82.6 & 30.0 & 8.3 & 3.44 & 62 & 1.5 & 40 \\
Claude Opus 5.5 / DSH & 43.71 & 93.0 & 69.1 & 58.7 & 11.10 & 52 & 1.7 & 7 \\
Kimi K3 / Kimi hosted & 82.10 & 72.2 & 13.5 & 5.2 & 4.10 & 62 & 3.2 & 23 \\
DeepSeek V4.1 Flash / WorkBuddy & 93.21 & 28.7 & 3.5 & 0.4 & 2.79 & 52 & 18.3 & 47 \\
DeepSeek V4.1 Flash / DSH & 76.16 & 72.6 & 30.4 & 20.9 & 7.73 & 60 & 3.4 & 10 \\
DeepSeek V4.1 Flash / Pi & 96.54 & 50.0 & 0.9 & 0.4 & 2.03 & 48 & 29.5 & 13 \\
DuMate / DuMate CLI & 12.21 & 97.4 & 90.4 & 59.1 & 23.89 & 38 & 3.7 & 20 \\
Seed-2.1-pro / Trae Work & 0.96 & 98.7 & 100.0 & 88.3 & 27.28 & 39 & 2.9 & 0 \\

\bottomrule
\end{tabular}}
\end{table}

\begin{figure}[!htbp]
\centering
\begin{minipage}[t]{0.49\textwidth}
\vspace{0pt}
\centering
\includegraphics[width=\linewidth]{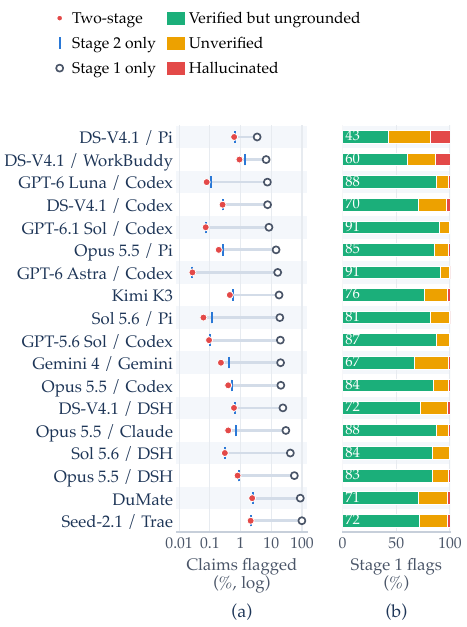}
\captionsetup{skip=0pt}
\caption{Comparison using saved audit verdicts for 18 configurations, ordered by the Stage 1 rate. (a) Share of claims each decision rule would flag (log scale): Stage 1 alone, Stage 2 alone, and the full two-stage audit. (b) Composition of Stage 1 flags; numbers give the verified-but-ungrounded percentage. No new audit judgments are made.}
\label{fig:ablation}
\end{minipage}\hfill
\begin{minipage}[t]{0.49\textwidth}
\vspace{0pt}
\centering
\includegraphics[width=\linewidth]{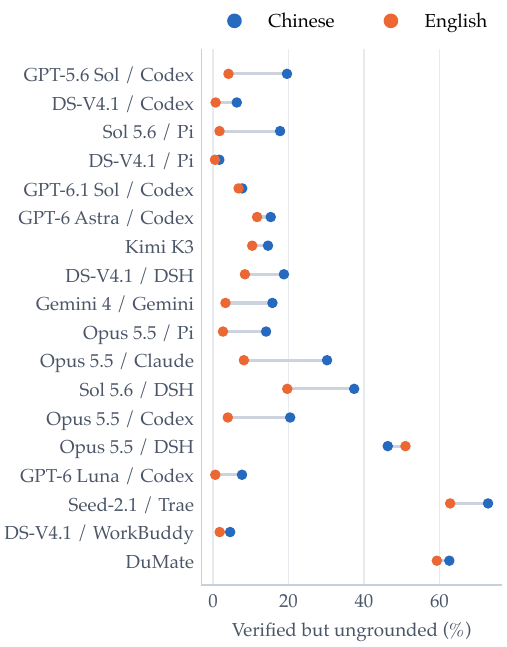}
\caption{VU rate on Chinese versus English questions. Rows follow \cref{tab:main}; VU denotes claims that are verified but ungrounded.}
\label{fig:language-vu}
\end{minipage}
\end{figure}




\subsection{Exploratory Classification of Reasoning Gaps}
\label{app:reasoning-gaps}
We secondarily classify all 1,960 distinct criteria in the explanation-and-reasoning dimension of the frozen rubrics, covering 230 questions and 18 configurations. Each criterion receives one primary category based on its wording, shared across all configurations regardless of satisfaction. Deterministic initial labels were reviewed criterion by criterion by an AI assistant, with 702 assignments revised; this is exploratory coding without independent human double-coding. No responses were regenerated or judgments rescored. These categories do not include criteria from the separate boundary-and-applicability dimension.

\begin{table}[!htbp]
\centering
\caption{Reasoning-category coverage and unmet rates. Each distinct criterion is judged for all 18 configurations; the unmet rate uses applicable judgments within its category as the denominator.}
\label{tab:reasoning-categories}
\small
\begin{tabularx}{\textwidth}{@{}Xrrrr@{}}
\toprule
Category & Criteria & \makecell{Applicable\\judgments} & Unmet & \makecell{Unmet rate\\(\%)} \\
\midrule
Explaining financial impacts & 836 & 15,048 & 3,628 & 24.11 \\
Comparison and synthesis & 376 & 6,768 & 1,351 & 19.96 \\
Scenarios and dynamic validation & 237 & 4,266 & 798 & 18.71 \\
Valuation and market expectations & 188 & 3,384 & 764 & 22.58 \\
Business, product, and institutional mechanisms & 164 & 2,952 & 654 & 22.15 \\
Investment strategy and trade-offs & 100 & 1,800 & 428 & 23.78 \\
Quantitative modeling and derivation & 59 & 1,062 & 293 & 27.59 \\
\midrule
Total & 1,960 & 35,280 & 7,916 & 22.44 \\
\bottomrule
\end{tabularx}
\end{table}

\paratitle{Explaining financial impacts.}Channels linking macroeconomic, policy, industry, and operating drivers to demand, revenue, costs, profits, cash flow, or prices.

\paratitle{Comparison and synthesis.}Distinguishing concepts and measurement scopes, establishing comparability, and integrating multiple factors into conclusions.

\paratitle{Scenarios and dynamic validation.}Conditional scenarios, time horizons and lags, risk triggers, follow-up validation, and limits to causal inference or extrapolation.

\paratitle{Valuation and market expectations.}Valuation methods, normalized earnings, implied expectations, margins of safety, and what prices already reflect.

\paratitle{Business, product, and institutional mechanisms.}Operating mechanisms of businesses, financial products, trading, institutions, and task-relevant technologies.

\paratitle{Investment strategy and trade-offs.}Investment objectives and constraints, allocation, hedging, execution, and risk--return or cost trade-offs.

\paratitle{Quantitative modeling and derivation.}Formulas, weighting, metric decomposition, currency and real/nominal conversions, compounding, and valuation-model construction.

\Cref{fig:reasoning-gaps}(a) reports each category's share of all 7,916 unmet judgments; panel (b) and \cref{tab:reasoning-categories} report unmet judgments divided by applicable judgments within each category. Shares therefore reflect both criterion frequency and non-fulfillment; they are not category-specific error probabilities or causal explanations of the L1--L4 score difference. An unmet criterion may indicate an omitted requirement rather than a factual error. Alternative assignment and split-allocation checks on 84 overlapping criteria retain transmission mechanisms and comparison/synthesis as the two largest categories by unmet count. These checks address category-boundary sensitivity, not independent validation of the taxonomy.

\subsection{Response Length and Claim Concentration}
\label{sec:rq-length}

This analysis uses unconfirmed claims, i.e., hallucinated plus unverified claims as in the unconfirmed rate $U$ (\cref{sec:score}). Response-level and claim-level views can rank configurations differently, so both should be reported (\cref{fig:length}). When Kimi K3's responses are grouped into quartiles by claim count, the share containing an unconfirmed claim rises from 43\% to 81\%, while the claim-level share stays between 3.6\% and 4.6\%. In every configuration, the 10\% most affected questions hold 38\%--93\% of the unconfirmed claims, so review can target specific responses. The association with length is not necessarily causal.

\begin{figure}[!t]
\centering
\includegraphics[width=\textwidth]{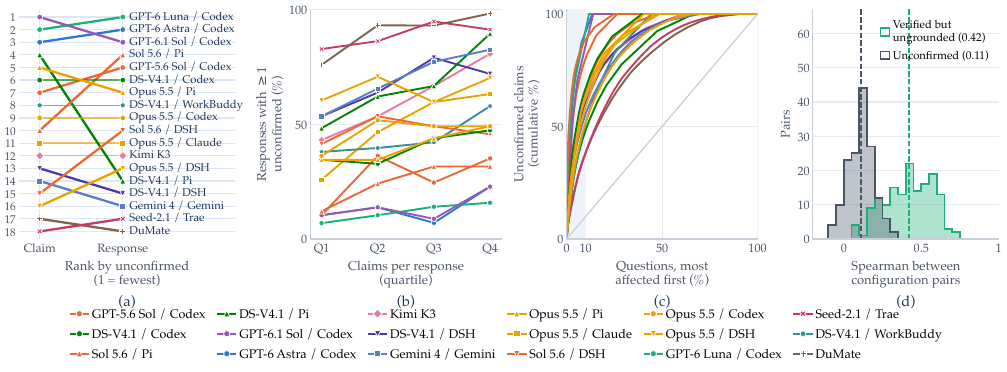}
\caption{Length, concentration, and question-level stability for 18 configurations. (a) Rank by unconfirmed claims at claim and response levels. (b) Share of responses with an unconfirmed claim by quartile of claims per response. (c) Concentration curves, with questions ordered by unconfirmed-claim count within each configuration. (d) Pairwise Spearman correlations of per-question VU and unconfirmed rates; dashed lines mark the means.}
\label{fig:length}
\end{figure}

\subsection{Breakdown by Complexity and Language}
\label{app:rq-task-breakdowns}

\paratitle{Complexity levels.}\Cref{fig:openness} decomposes the audit outcomes by the complexity levels of \cref{tab:task-complexity} on the 230 questions, which contain 10, 70, 86, and 64 questions at L1--L4. Simpler questions are less often grounded in the trace: averaged over the 18 configurations, the VU rate is 36.1\% and 36.0\% at L1 and L2 but 15.9\% and 13.3\% at L3 and L4, and it is higher on L1--L2 than on L3--L4 in all 18 configurations. Systems thus often answer simpler questions with claims that are verified but ungrounded, so the VU rate is highest where hallucinations are least frequent (\cref{sec:rq-difficulty}). Claim-level shares pool the claims within each level, and the breakdown is descriptive.

\begin{figure}[!t]
\centering
\begin{minipage}[c]{0.5\textwidth}
\includegraphics[width=\linewidth]{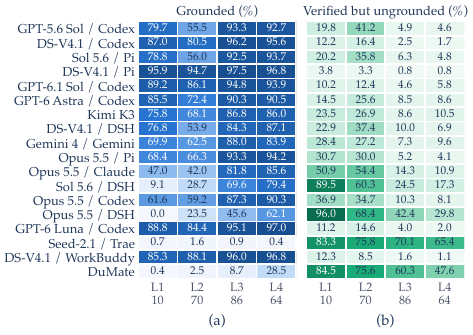}
\end{minipage}\hfill
\begin{minipage}[c]{0.46\textwidth}
\captionsetup{skip=0pt}
\caption{Audit outcomes by complexity level on the 230 questions, rows as in \cref{fig:fulfillment}. (a) Grounded rate. (b) VU rate. Claims are pooled within each level, and the column labels give the number of questions. Responses with a hallucinated claim by level are shown in \cref{fig:difficulty}(a).}
\label{fig:openness}
\end{minipage}
\end{figure}

\paratitle{Language.}In 17 of the 18 configurations, the VU rate is higher on Chinese questions than on English questions (\cref{fig:language-vu}). The exception is Claude Opus 5.5 on DSH, at 46.3\% for Chinese and 51.0\% for English. For example, the Claude Opus 5.5 configuration on Claude Code CLI is at 30.2\% versus 8.2\%, and GPT-5.6 Sol on Codex CLI at 19.6\% versus 4.1\%.

\end{document}